\documentclass[pdflatex,sn-mathphys-num]{sn-jnl}

\usepackage{graphicx}%
\usepackage{multirow}%
\usepackage{amsmath,amssymb,amsfonts}%
\usepackage{amsthm}%
\usepackage{mathrsfs}%
\usepackage[title]{appendix}%
\usepackage{xcolor}%
\usepackage{textcomp}%
\usepackage{manyfoot}%
\usepackage{booktabs}%
\usepackage{algorithm}%
\usepackage{algorithmicx}%
\usepackage{algpseudocode}%
\usepackage{listings}%

\usepackage{url}
\usepackage{hyperref}

\theoremstyle{thmstyleone}%
\theoremstyle{thmstyletwo}%

\theoremstyle{thmstylethree}%

\begin{document}

\title[Article Title]{Label-Efficient Learning for Ground-Based Sky-Image Classification: A Benchmark of Transfer Learning, Active Learning, and Pseudo-Labeling on GCD}


\author[1]{\fnm{Esther} \sur{Bou Dagher}}\email{esther.bou-dagher@dauphine.psl.eu}

\author*[2]{\fnm{Viktoriya} \sur{Bu-Dager}}\email{Viktoriya.Bu-Dager@aru.ac.uk}

\author*[3]{\fnm{Boguslaw} \sur{Zegarlinski}}\email{bzegarlinski@impan.pl}

\affil[1]{\orgdiv{CEREMADE}, \orgname{Universit{\'e} Paris Dauphine-PSL}, \orgaddress{\city{Paris}, \country{France}}}

\affil*[2]{\orgdiv{School of Computing and Information Science}, \orgname{Anglia Ruskin University}, \orgaddress{\city{Cambridge}, \country{United Kingdom}}}

\affil*[3]{\orgdiv{Institute of Mathematics}, \orgname{Polish Academy of Sciences}, \orgaddress{\city{Warsaw}, \country{Poland}}}


\abstract{Accurate ground-based cloud classification is important for atmospheric monitoring, solar-energy forecasting, aviation weather assessment, and climate observation systems. However, reliable sky-image annotation is time-consuming, especially when cloud types are visually similar or mixed. In this work, we study the label efficiency of deep learning for ground-based cloud classification using the Ground-based Cloud Dataset (GCD). Rather than proposing a new architecture, we benchmark three practical learning strategies under limited annotation budgets: supervised transfer learning, uncertainty-based active learning, and high-confidence pseudo-labeling. An ImageNet-pretrained ResNet50 is used as a common frozen backbone, and experiments are repeated over five random seeds for label budgets ranging from $1\%$ to $100\%$ of the training labels. Supervised transfer learning is already highly label-efficient: test accuracy increases from $0.635 \pm 0.018$ with $1\%$ labels to  $0.730 \pm 0.002$ with $40\%$ labels, approaching the full-label performance of $0.735 \pm 0.003$. Active learning and pseudo-labeling are competitive with supervised sampling and provide small improvements for some metrics and budgets, but neither produces a large or consistent aggregate gain. Diagnostic analyses show that accepted pseudo-labels are highly reliable, with accuracy ranging from $0.946$ to $0.977$, but are biased toward easier, high-confidence sky-type groups. In contrast, uncertainty sampling preferentially queries uncertain samples from visually challenging groups, including Mixed and the visually confusable Stratocumulus and Cumulonimbus groups, but these targeted acquisitions yield only modest performance improvements. Overall, these results show that transfer learning can substantially reduce annotation requirements for GCD, while simple active and semi-supervised strategies provide limited additional gains over a strong supervised baseline.}




\maketitle

\section{Introduction}\label{sec:introduction}
Cloud observation is essential for weather monitoring, climate research, solar energy forecasting, aviation weather assessment, and local environmental observation. Cloud type, structure, and temporal evolution influence radiative transfer, precipitation processes, surface irradiance, and short-term atmospheric conditions \citep{zhangCloudNetGroundBasedCloud2018, lvClassificationGroundBasedCloud2022}. Ground-based sky imaging complements satellite observations by providing local, high-resolution views of cloud fields from below. Such images are therefore useful for operational tasks such as sky-condition assessment, cloud-type classification, and short-term forecasting.

Automatic ground-based cloud classification remains a challenging task. Cloud categories are often difficult to distinguish, particularly when cloud fields contain multiple cloud types or exhibit gradual transitions between morphologies. Some classes, such as clear sky, are relatively easy to identify, whereas others, including altocumulus, stratocumulus, cumulonimbus, and mixed cloud scenes, can share similar texture, illumination, scale, and shape. Early cloud classification approaches relied on hand-crafted features, such as colour, texture, and statistical descriptors. More recently, deep learning has significantly improved image-based cloud classification by learning features directly from data, and convolutional neural networks and related architectures have become widely used for this task \citep{liCloudDenseNetLightweightGroundBased2023, lvClassificationGroundBasedCloud2022}.

Most recent work on ground-based cloud classification has focused on model design. Proposed methods include lightweight convolutional networks, graph-based models, attention mechanisms, transformer-inspired architectures, and hybrid feature-fusion approaches \citep{liCloudDenseNetLightweightGroundBased2023, shiImprovedRepVGGGroundbased2024, liNovelMethodGroundBased2022, liuGroundBasedCloudClassification2020}. These studies show that specialised deep networks can achieve strong classification performance on public cloud-image datasets. However, practical deployment depends not only on architecture, but also on annotation cost. In many observational settings, labelled cloud images are expensive to obtain because reliable annotation requires domain expertise, consistent interpretation across cloud categories, and careful handling of ambiguous or mixed cases. Therefore, an important practical question is not only which architecture achieves the highest performance with fully labelled data, but also how much labelled data is required to achieve useful classification performance.

Label-efficient learning addresses this question by reducing dependence on manually labelled data. In remote sensing, label scarcity has motivated transfer learning, self-supervised learning, semi-supervised learning, and active learning approaches \citep{wangSelfSupervisedLearningRemote2022}. In ground-based cloud classification, contrastive self-supervised learning has been proposed as a way to reduce dependence on labelled data \citep{lvClassificationGroundBasedCloud2022}. Pseudo-labeling provides another simple semi-supervised strategy: a model trained on a small labelled subset assigns labels to high-confidence unlabelled samples, which are then used for further training. Active learning takes a different approach by selecting unlabelled samples that are expected to provide the greatest information gain, such as those for which model predictions are most uncertain. Both strategies are attractive in cloud-image analysis, where large collections of unlabelled sky images are often available while expert-labelled data remain limited.

Despite this motivation, there is still a need for a systematic empirical benchmark of label-efficient methods for ground-based cloud classification. In particular, it is important to compare active learning and pseudo-labeling against a strong supervised transfer-learning baseline under the same label budgets. Such a benchmark should evaluate not only overall classification accuracy, but also class-balanced performance metrics, variability across random seeds, and diagnostic analyses that explain why label-efficient strategies do or do not improve performance. This is especially important for imbalanced cloud datasets, where high overall accuracy can hide poor performance on minority or visually ambiguous classes.

In this study, we focus on the practical problem of reducing annotation requirements for ground-based cloud classification. Using the Ground-based Cloud Dataset (GCD), we evaluate whether simple label-efficient learning strategies can improve upon a strong supervised transfer-learning baseline. Rather than proposing a new neural-network architecture, the aim is to provide a controlled benchmark and diagnostic analysis of when active and semi-supervised strategies are useful in this setting.

The main contributions of this work are as follows:
\begin{itemize}
\item We quantify the supervised label-efficiency curve of frozen ImageNet-pretrained ResNet50 on GCD, showing how performance changes as the labelled training budget increases.
\item We compare supervised stratified random sampling, uncertainty-based active learning, and high-confidence pseudo-labeling under matched label budgets, a fixed training protocol, and five random seeds.
\item We evaluate performance using overall and class-balanced metrics, including accuracy, macro-F1 score, balanced accuracy, weighted-F1 score, and per-class F1 scores.
\item We analyse pseudo-label coverage, accepted accuracy, rejected prediction accuracy, confidence, and class-relative selection within the unlabelled training pool.
\item We examine active-learning query enrichment relative to the remaining unlabelled pool to determine whether uncertainty sampling preferentially selects visually ambiguous cloud categories.
\end{itemize}

\section{Related work}\label{sec:related_work}

\subsection{Ground-based cloud image classification}

Ground-based cloud classification has been studied using both traditional image-processing methods and modern deep-learning approaches. Early methods relied on hand-crafted descriptors designed to capture colour, texture, shape, and local image structure. Representative examples include feature extraction from whole-sky images, automatic whole-sky cloud classification, block-based statistical and texture descriptors, weighted local binary patterns, and texton-based representations\citep{calboFeatureExtractionWholeSky2008, heinleAutomaticCloudClassification2010, chengBlockbasedCloudClassification2015, liuGroundbasedCloudClassification2015, devCategorizationCloudImage2015}. Later descriptor-based approaches used region covariance descriptors and Riemannian bag-of-features representations to improve ground-based cloud classification \citep{tangImprovingCloudType2021}. These methods provided important baselines, but their performance depends strongly on the quality of the manually designed representation and can be limited by the large intra-class variability of cloud images.

Deep learning has substantially changed the methodology for ground-based cloud classification by allowing feature representations to be learned directly from images. DeepCloud showed that convolutional visual features extracted from convolutional neural networks can improve ground-based cloud image categorization compared with traditional descriptors \citep{yeDeepCloudGroundBasedCloud2017}. CloudNet proposed a dedicated convolutional neural-network for meteorological cloud classification and introduced the Cirrus Cumulus Stratus Nimbus (CCSN) dataset with cloud categories under meteorological standards \citep{zhangCloudNetGroundBasedCloud2018}. Subsequent studies developed more specialised architectures, including graph convolutional models, heterogeneous feature-learning methods, attention-based networks, transformer-inspired models, and hybrid CNN--transformer approaches \citep{liuGroundBasedCloudClassification2020, liuGroundBasedRemoteSensing2022, liNovelMethodGroundBased2022}.

Recent work has also focused on models designed for larger public ground-based cloud datasets. CloudDenseNet proposed a lightweight reconstructed DenseNet architecture and evaluated it on large-scale ground-based cloud datasets, including GCD \citep{liCloudDenseNetLightweightGroundBased2023}. Improved RepVGG-style models with attention mechanisms have similarly been proposed to capture both local texture and longer-range spatial structure in cloud images \citep{shiImprovedRepVGGGroundbased2024}. These architecture-focused studies demonstrate the strong performance of deep networks for cloud classification. However, they typically evaluate performance using the available labelled training set, whereas the question of how performance changes under systematically reduced label budgets has received less attention. 

\subsection{Label scarcity and label-efficient learning in remote sensing}

Label scarcity is a common challenge in remote sensing and environmental image analysis. Labelling often requires expert knowledge, quality control, and consistency across sensors, locations, and acquisition conditions. In ground-based cloud classification, these difficulties are amplified by mixed cloud fields, ambiguous visual boundaries, and gradual transitions between cloud types. Label-efficient learning is therefore important for deploying cloud-classification systems in new observational settings.

Several label-efficient learning paradigms have been explored in remote sensing, including transfer learning, self-supervised learning, semi-supervised learning, and active learning. Transfer learning is particularly attractive when the target dataset is not large enough to train a deep model from scratch. A model pretrained on a large image dataset can provide generic visual features, while the final layers are adapted to the target task. In this work, we use ResNet50 as the shared backbone, following the residual-learning framework introduced by \citet{heDeepResidualLearning2016}, and initialise the model with ImageNet-pretrained weights, following the transfer-learning paradigm enabled by large-scale datasets such as ImageNet \citep{dengImageNetLargescaleHierarchical2009}. This provides a strong supervised baseline against which active learning and pseudo-labeling can be evaluated.

Self-supervised learning is another strategy for reducing dependence on labels. In remote sensing, self-supervised approaches have been studied as a way to exploit large unlabelled image archives \citep{wangSelfSupervisedLearningRemote2022}. In ground-based cloud classification, contrastive self-supervised learning has been used to learn cloud-image representations before supervised classification \citep{lvClassificationGroundBasedCloud2022}. Our study is related in motivation, but different in scope: instead of proposing a new representation-learning method, we benchmark simple and reproducible label-efficient strategies under matched annotation budgets.

\subsection{Active learning for remote sensing classification}

Active learning aims to reduce annotation cost by selecting informative unlabelled samples for labelling \citep{settlesActiveLearningLiterature2009}. In remote sensing, active learning has long been studied because labelled data are expensive and random sampling may not capture the full spatial, spectral, or visual diversity of the data. Common strategies include uncertainty sampling, margin sampling, entropy-based criteria, query-by-committee, and diversity-aware selection. Surveys of active learning in remote sensing emphasise that training-set quality is critical for classification performance and that uncertainty-based heuristics provide simple and practical sample-selection rules \citep{tuiaSurveyActiveLearning2011}.

Deep active learning extends these ideas to neural networks and large image archives. Recent work in remote-sensing image classification has considered single-label, multi-class, and multi-label settings, often combining uncertainty with diversity to avoid selecting redundant samples \citep{mollenbrokDeepActiveLearning2023}. These studies motivate active learning as a practical strategy for reducing annotation cost in remote-sensing classification. In the present study, we examine whether this strategy provides additional benefit over a strong transfer-learning baseline for ground-based cloud classification.

\subsection{Pseudo-labeling and semi-supervised learning}

Semi-supervised learning uses both labelled and unlabelled data during training. Pseudo-labeling is one of the simplest semi-supervised strategies: a model trained on labelled data predicts labels for unlabelled samples, and high-confidence predictions are treated as additional training labels \citep{lee2013pseudo}. This approach is attractive because it is easy to implement and can be applied to standard supervised architectures. However, its effectiveness depends strongly on pseudo-label quality. Incorrect pseudo-labels can reinforce model errors, while overly conservative thresholds may select mostly easy examples and add little new information.

Pseudo-labeling and related semi-supervised methods have been studied in remote sensing, especially for scene classification and semantic segmentation, where annotation costs are high \citep{wangSemiSupervisedSemanticSegmentation2022, fengImprovingSemisupervisedRemote2025, ranPseudoLabelingMethods2025}. These studies highlight both the potential and the limitations of pseudo-label-based learning. Reliable pseudo-labels can improve performance, but noisy or class-biased pseudo-labels may limit gains. Because pseudo-labeling can be affected by confirmation errors and class bias, diagnostic analyses of pseudo-label quality are important when evaluating semi-supervised methods.

\subsection{Positioning of this work}

The existing literature shows that deep learning can achieve strong ground-based cloud classification performance, and that label-efficient learning is important in remote sensing. However, architecture-focused cloud-classification studies do not directly answer how much labelled data is needed, or whether simple active and semi-supervised strategies improve over a strong transfer-learning baseline. This study addresses that gap by focusing on annotation efficiency rather than architectural novelty.

\section{Materials and methods}\label{sec:methods}

\subsection{Dataset and data partitioning}\label{sec:dataset}

Experiments were conducted using the publicly available Ground-based Cloud Dataset, which contains seven sky-type groups defined according to the International Cloud Classification System and practical visual similarity. The seven GCD groups are: Cumulus; Altocumulus/Cirrocumulus; Cirrus/Cirrostratus; Clear sky; Stratocumulus/Stratus/Altostratus; Cumulonimbus/Nimbostratus; and Mixed cloud. Images with cloudiness no greater than 10\% are included in the Clear sky group \citep{liuGroundBasedRemoteSensing2022}. For readability, we use shortened labels throughout the manuscript: Cumulus, Altocumulus, Cirrus, Clear sky, Stratocumulus, Cumulonimbus, and Mixed. These shortened labels refer to the corresponding GCD sky-type groups rather than to individual cloud types. We use this convention consistently in figures, tables, and the subsequent analysis. 

\begin{figure}[!t]
    \centering
    \includegraphics[width=\linewidth]{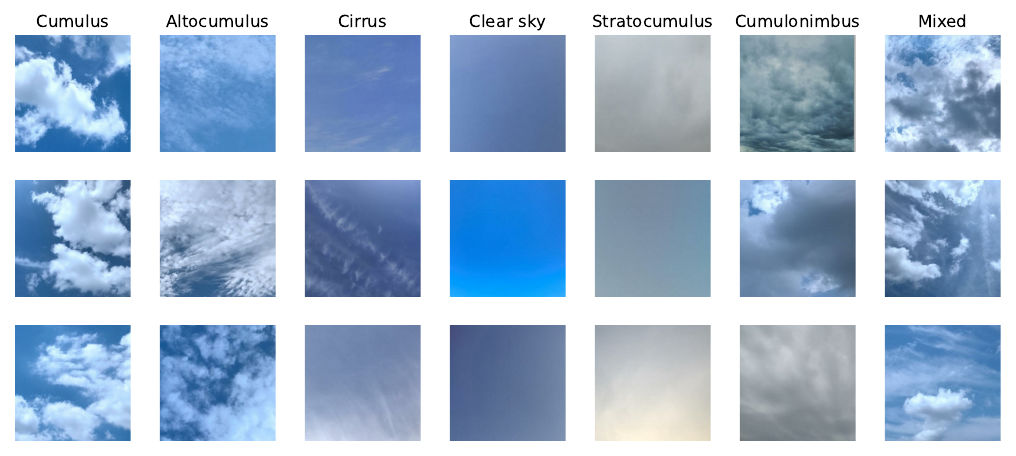}
    \caption{\textbf{Representative training images from the seven GCD sky-type groups.} Shortened labels are used for readability and refer to the corresponding GCD sky-type groups. The examples illustrate both visually distinctive groups, such as Clear sky, and more visually ambiguous groups, such as Mixed, Stratocumulus, and Cumulonimbus.}
    \label{fig:gcd_examples}
\end{figure}

The dataset comprises a 10,000-image training set and an independent 9,000-image test set. Representative examples from each GCD sky-type group are shown in Figure~\ref{fig:gcd_examples}.

In the final benchmark, no labelled validation subset was used for model selection. Instead, label-budget subsets were sampled from the full training set, and fixed training schedules were used across all methods, label budgets, and random seeds. Thus, the reported label budgets correspond directly to the labelled images used for model optimisation, without an additional labelled validation resource. The independent test set was loaded without shuffling and was used exclusively for final performance evaluation. It was excluded from training, pseudo-label generation, active-learning sample acquisition, hyperparameter selection, and all other model-development steps.

The dataset exhibits substantial class imbalance. Within the training set, the Cumulonimbus group was the largest (3,003 images), whereas the Mixed group was the smallest (348 images). A similar distribution was observed in the independent test set. The complete distribution of the seven GCD sky-type groups is presented in Table~\ref{tab:class_distribution}.

\begin{table}[htbp]
\caption{\textbf{Distribution of the seven GCD sky-type groups in the training and independent test sets.} Label-budget subsets were sampled from the full 10,000-image training set, with no additional labelled validation subset used for model selection.}
\label{tab:class_distribution}

\begin{tabular*}{\textwidth}{@{}p{\textwidth}@{}}
\centering

\begin{tabular}{@{}lrr@{}}
\toprule
Sky-type group & Training & Test \\
\midrule
Cumulus & 775 & 750 \\
Altocumulus & 725 & 750 \\
Cirrus & 1,153 & 753 \\
Clear sky & 2,150 & 1,589 \\
Stratocumulus & 1,846 & 1,790 \\
Cumulonimbus & 3,003 & 2,761 \\
Mixed & 348 & 607 \\
\midrule
Total & 10,000 & 9,000 \\
\bottomrule
\end{tabular}

\end{tabular*}

\vspace{8pt}

\footnotesize
\emph{Note:} Shortened labels correspond to the GCD sky-type groups defined in Section~\ref{sec:dataset}.

\end{table}

The class imbalance motivated the use of class-sensitive evaluation metrics in addition to overall accuracy. Specifically, macro-F1 score and balanced accuracy were included to ensure that performance on majority groups did not obscure weaker performance on minority or visually ambiguous cloud categories.

\subsection{Image preprocessing and data pipeline}\label{sec:preprocessing}

All images were loaded using the TensorFlow/Keras image data pipeline. Images were resized to $224 \times 224$ pixels, corresponding to the standard input resolution used by ResNet50. Before being passed to the network, images were converted to 32-bit floating-point tensors and preprocessed by the standard ResNet50 preprocessing function used for the ImageNet-pretrained backbone. 

For each seed and label budget, the selected labelled indices were used to construct a dataset from the preprocessed training set. This resulting dataset was batched with a mini-batch size of $32$ and prefetched using TensorFlow automatic tuning.

The same preprocessing pipeline was applied to all training and test images, ensuring that differences between the supervised, active-learning, and pseudo-labeling experiments arose from the label-selection and training protocols rather than from differences in input preparation.

No data augmentation was applied in the main experiments. This provided a controlled comparison of supervised transfer learning, active learning, and pseudo-labeling under identical preprocessing and modelling conditions. Consequently, differences in performance reflect the learning strategies and label availability rather than augmentation effects.

\subsection{Transfer-learning backbone}
\label{sec:backbone}

All experiments used the same transfer-learning architecture. An ImageNet-pretrained ResNet50 was selected as the backbone because it provides a strong and widely used visual feature extractor, making it well suited for benchmarking label-efficient learning strategies. The network was loaded without its original classification head using \texttt{include\_top=False}, and a task-specific classification head was attached. The ResNet50 backbone was kept frozen throughout the benchmark using \texttt{base\_model.trainable = False}, so that the comparisons focused on labelled-data availability and learning strategy rather than backbone fine-tuning.

Given a preprocessed input image $\boldsymbol{x}$, the ResNet50 backbone produced a spatial feature representation,
\begin{equation}
\boldsymbol{z}
=
f_{\mathrm{ResNet50}}(\boldsymbol{x}).
\end{equation}
The resulting feature map was processed by a global average pooling (GAP) layer, followed by a dense layer with $256$ units and ReLU activation, dropout ($0.5$), and a final seven-class softmax output layer. The resulting architecture can be summarised as
\begin{equation}
\begin{aligned}
\boldsymbol{x}
&\longmapsto
f_{\mathrm{ResNet50}}(\boldsymbol{x})
\longmapsto
\mathrm{GAP}
\\
&\longmapsto
\mathrm{Dense}(256)+\mathrm{ReLU}
\longmapsto
\mathrm{Dropout}(0.5)
\longmapsto
\mathrm{Softmax}(7).
\end{aligned}
\end{equation}

All models were trained using sparse categorical cross-entropy loss and the Adam optimiser. To ensure that the reported label budgets did not depend on an additional labelled validation resource, no validation-loss-based early stopping, validation-based model checkpointing, or validation-based learning-rate scheduling was used. Instead, models were trained for fixed schedules, with the model obtained at the final training epoch evaluated on the independent test set. Class weights were not used; class imbalance was instead accounted for through stratified label sampling and class-sensitive evaluation metrics.

The same backbone architecture and preprocessing pipeline were used across the supervised, active-learning, and pseudo-labeling experiments. Method-specific training schedules, including learning rates and numbers of epochs, are described in the corresponding subsections.

\subsection{Label-budget protocol and supervised baseline}\label{sec:label_budget_supervised}

To quantify the effect of annotation availability, models were trained using label budgets of $1\%$, $3\%$, $5\%$, $10\%$, $20\%$, $40\%$, and $100\%$ of the 10,000-image training set. For each budget, labelled samples were selected by class-stratified random sampling without replacement, retaining at least one image from each class whenever possible. This produced labelled sets of $100$, $299$, $500$, $1000$, $2001$, $3999$, and $10000$ images, respectively; small deviations from the nominal percentages arose from class-wise rounding. The supervised baseline used only the labelled subset at each budget.

The sampling procedure was repeated for five random seeds, $0$, $1$, $2$, $3$, and $4$. For each seed and label budget, a fresh ImageNet-pretrained ResNet50 model with a randomly initialised classification head was trained using the architecture and optimisation framework described in Section~\ref{sec:backbone} for a fixed schedule of $10$ epochs. The training subset was shuffled during optimisation, while the full training set was loaded in a fixed order to ensure reproducible index-based sampling. The independent test set was not used for training, sample selection, pseudo-label generation, active-learning acquisition, or model selection, and was used only for final evaluation. The $100\%$-label model served as the reference baseline for comparison with the reduced-label supervised, active-learning, and pseudo-labeling experiments.

\subsection{Uncertainty-based active learning}\label{sec:active_learning}

To evaluate whether annotation efficiency could be improved by selecting informative samples, we implemented a pool-based active-learning strategy using uncertainty sampling. Active learning was evaluated at $1\%$, $3\%$, $5\%$, $10\%$, $20\%$, and $40\%$ label budgets. The $100\%$ budget was excluded because no unlabelled samples remained for acquisition.

For each random seed, active learning was initialised with the same class-stratified $1\%$ labelled subset used in the supervised experiments, containing $100$ images. The remaining training images formed the initial unlabelled pool. 
A model was first trained on the initial labelled subset using the fixed supervised protocol described in Section~\ref{sec:backbone}. It was then applied to the current unlabelled pool to compute class probabilities for each unlabelled image.

For an unlabelled image $\boldsymbol{x}$, the uncertainty score was defined as
\begin{equation}
u(\boldsymbol{x})
=
1-\max_{c} p_\theta(y=c \mid \boldsymbol{x}),
\end{equation}
where $p_\theta(y=c \mid \boldsymbol{x})$ is the predicted probability of class $c$ under the current model parameters $\theta$. Larger values of $u(\boldsymbol{x})$ correspond to lower maximum predicted confidence and therefore higher uncertainty.

At each acquisition step, the most uncertain samples were selected from the current unlabelled pool, their ground-truth labels were revealed, and they were added to the labelled training set. The labelled set was expanded according to the sequence
\begin{equation}
1\% \rightarrow 3\% \rightarrow 5\% \rightarrow 10\% \rightarrow 20\% \rightarrow 40\%.\label{eq:active_label_budgets}
\end{equation}
The resulting labelled-set sizes were $100$, $299$, $500$, $1000$, $2001$, and $3999$ images, respectively, matching the rounded label-budget sizes used in the supervised experiments.

After each acquisition step, the model was warm-started from the previous active-learning stage and further trained on the enlarged labelled set. The initial $1\%$ model was trained for $10$ epochs using Adam with learning rate $10^{-3}$. Subsequent acquisition stages were trained for a fixed $5$ epochs using Adam with learning rate $5\times10^{-4}$. After each stage, the final model was evaluated on the independent test set.

\subsection{High-confidence pseudo-labeling}\label{sec:pseudo_labeling}

To evaluate a simple semi-supervised learning strategy, we implemented high-confidence pseudo-labeling at label budgets of $1\%$, $3\%$, $5\%$, $10\%$, $20\%$, and $40\%$ of the 10,000-image training set. The $100\%$ budget was excluded because no unlabelled training samples remain for pseudo-label generation.

For each random seed and label budget, a class-stratified labelled subset was sampled from the training set, while the remaining images formed the unlabelled pool. A first-stage supervised model was trained for $10$ epochs with a learning rate of $10^{-3}$ on the labelled subset only, using the fixed training protocol described in Section~\ref{sec:backbone}. This model was then used to predict class probabilities for all images in the unlabelled pool.

For an unlabelled image $\boldsymbol{x}$, the predicted pseudo-label $\widehat{y}$ and confidence $q(\boldsymbol{x})$ were defined as
\begin{equation}
\widehat{y}
=
\arg\max_{c} p_{\theta}(y=c \mid \boldsymbol{x}),
\qquad
q(\boldsymbol{x})
=
\max_{c} p_{\theta}(y=c \mid \boldsymbol{x}).
\end{equation}

Only predictions satisfying
\begin{equation}
q(\boldsymbol{x}) \geq \tau,
\qquad
\tau = 0.95,
\end{equation}
were retained as pseudo-labels.

The accepted pseudo-labelled images were combined with the original labelled subset and shuffled together to form second-stage training set. The second-stage model was initialised from the first-stage checkpoint and trained for a fixed $5$ additional epochs using Adam with learning rate $5\times 10^{-4}$. The number of optimiser updates in the second stage was determined by the size of the combined labelled and pseudo-labelled dataset.

To separate the effect of pseudo-label information from the effect of additional optimisation, we also trained a continued-supervised control for each seed and label budget. This control was initialised from the same first-stage checkpoint as the semi-supervised model and trained using the same learning rate and the same total number of optimiser updates, but using only the original labelled subset. Thus, the semi-supervised model and the continued-supervised control differed only in whether the second-stage updates used pseudo-labelled samples in addition to the labelled data.

The first-stage model, the continued-supervised control, and the semi-supervised model were all evaluated on the independent test set. The main pseudo-labeling effect was measured by comparing the semi-supervised model with the update-matched continued-supervised control. Pseudo-label coverage and mean confidence were also recorded to characterise the size and confidence of the accepted pseudo-labelled set.

\subsection{Evaluation metrics}\label{sec:evaluation_metrics}

Model performance was evaluated on the independent 9,000-image test set. For each method, label budget, and random seed, we computed accuracy, balanced accuracy, macro-F1, weighted-F1, per-class metrics, and the confusion matrix. Results are reported as the mean and standard deviation over five random seeds.

Accuracy was defined as the proportion of correctly classified test images:
\begin{equation}
\mathrm{Accuracy}
=
\frac{1}{N}
\sum_{i=1}^{N}
\mathbb{I}\left(\widehat{y}_i = y_i\right),
\end{equation}
where $N$ is the number of test images, $y_i$ is the true class label, $\widehat{y}_i$ is the predicted class label, and $\mathbb{I}(\cdot)$ is the indicator function.

Because the GCD dataset is imbalanced across sky-type groups, we also used class-balanced metrics. For each class $c$, precision, recall, and F1-score were computed from the corresponding true positives ($\mathrm{TP}_c$), false positives ($\mathrm{FP}_c$), and false negatives ($\mathrm{FN}_c$):
\begin{equation}
\begin{array}{lcl}
\mathrm{Precision}_c
& = &
\frac{\mathrm{TP}_c}{\mathrm{TP}_c+\mathrm{FP}_c},
\\[5pt]
\mathrm{Recall}_c
& = &
\frac{\mathrm{TP}_c}{\mathrm{TP}_c+\mathrm{FN}_c},
\\[5pt]
\mathrm{F1}_c
& = &
\frac{2\,\mathrm{Precision}_c\,\mathrm{Recall}_c}
{\mathrm{Precision}_c+\mathrm{Recall}_c}.
\end{array}
\end{equation}

Macro-averaged metrics assign equal weight to each of the $C=7$ classes. For example,
\begin{equation}
\mathrm{Macro\text{-}F1}
=
\frac{1}{C}
\sum_{c=1}^{C}
\mathrm{F1}_c.
\end{equation}
Balanced accuracy was defined as the mean recall across the seven sky-type groups:
\begin{equation}
\mathrm{Balanced\ accuracy}
=
\frac{1}{C}
\sum_{c=1}^{C}
\mathrm{Recall}_c.
\end{equation}
These metrics are important here because strong performance on frequent sky-type groups, such as Cumulonimbus or Clear sky, may otherwise hide weaker performance on minority or visually ambiguous groups.

Weighted-F1 was calculated by weighting each class-specific F1-score by its test-set support:
\begin{equation}
\mathrm{Weighted\text{-}F1}
=
\sum_{c=1}^{C}
\frac{N_c}{N}
\mathrm{F1}_c,
\end{equation}
where $N_c$ is the number of test samples belonging to class $c$.

Confusion matrices were row-normalised for visualisation, so that each entry represents the fraction of images from a given true sky-type group assigned to each predicted group. They were used to identify persistent confusions between visually similar or mixed cloud groups.

\subsection{Diagnostic analyses}\label{sec:diagnostics}

\subsubsection{Active-query enrichment}\label{sec:active_query_enrichment}

To interpret the behaviour of uncertainty sampling, we analysed the sky-type-group composition of the samples selected at each acquisition step. The acquisition strategy selected samples solely according to predictive uncertainty and did not use class labels. True sky-type-group labels were inspected only after acquisition for diagnostic purposes.

For each acquisition step from label budget $a$ to label budget $b$, let $\mathcal{U}^{(a)}$ denote the unlabelled pool immediately before acquisition, and let $U^{(a)} = |\mathcal{U}^{(a)}|$. Let $U_c^{(a)}$ denote the number of remaining unlabelled samples from sky-type group $c$ immediately before that acquisition step. Similarly, let $Q^{(a\rightarrow b)}$ denote the total number of queried samples, and let $Q_c^{(a\rightarrow b)}$ denote the number of queried samples from sky-type group $c$.

The queried fraction and the corresponding pre-acquisition pool fraction were defined as
\begin{equation}
r_c^{(a\rightarrow b)}
=
\frac{Q_c^{(a\rightarrow b)}}{Q^{(a\rightarrow b)}},
\qquad
\pi_c^{(a)}
=
\frac{U_c^{(a)}}{U^{(a)}}.
\end{equation}
Here, $\pi_c^{(a)}$ is computed using the remaining unlabelled pool immediately before the acquisition step, rather than the original full training set. This accounts for the fact that earlier acquisitions may deplete some sky-type groups from the unlabelled pool.

The active-query enrichment ratio was then calculated as
\begin{equation}
E_c^{(a\rightarrow b)}
=
\frac{
r_c^{(a\rightarrow b)}
}{
\pi_c^{(a)}
}.
\end{equation}
Values above one indicate that sky-type group $c$ was over-represented among queried samples relative to its availability in the current unlabelled pool, whereas values below one indicate under-representation. The analysis was performed for each active-learning acquisition step in~\eqref{eq:active_label_budgets} and summarised across five random seeds.

\subsubsection{Pseudo-label quality}\label{sec:pseudo_label_quality}

To interpret the behaviour of pseudo-labeling, we performed a post hoc diagnostic analysis of pseudo-label coverage, accuracy, confidence, and sky-type-group selection. Because GCD provides ground-truth labels for all training images, pseudo-label correctness could be evaluated after pseudo-label generation. These ground-truth labels were used only for diagnostic analysis and were not used during pseudo-label selection or second-stage training.

For each label budget and random seed, the first-stage supervised model was applied to the corresponding unlabelled training pool. Let $\mathcal{U}$ denote the unlabelled pool, with $U = |\mathcal{U}|$. For each image $i \in \mathcal{U}$, the model produced a predicted pseudo-label $\widehat{y}_i$ and a confidence score $q_i$, defined as the maximum predicted class probability. The accepted pseudo-label set was defined as
\begin{equation}
\mathcal{A}
=
\{i \in \mathcal{U}: q_i \geq \tau\},
\qquad
\tau = 0.95,
\end{equation}
with $A = |\mathcal{A}|$. The rejected set was defined as $\mathcal{R}=\mathcal{U}\setminus\mathcal{A}$.

Pseudo-label coverage and accepted pseudo-label accuracy were defined as
\begin{equation}
\mathrm{Coverage}
=
\frac{A}{U},
\qquad
\mathrm{Accepted\ accuracy}
=
\frac{1}{A}
\sum_{i \in \mathcal{A}}
\mathbb{I}\left(\widehat{y}_i = y_i\right),
\end{equation}
where $y_i$ is the ground-truth label used only for post hoc evaluation, and $\mathbb{I}(\cdot)$ is the indicator function. For comparison, we also computed the accuracy of rejected predictions,
\begin{equation}
\mathrm{Rejected\ accuracy}
=
\frac{1}{|\mathcal{R}|}
\sum_{i \in \mathcal{R}}
\mathbb{I}\left(\widehat{y}_i = y_i\right).
\end{equation}
Mean accepted confidence was computed as
\begin{equation}
\mathrm{Accepted\ confidence}
=
\frac{1}{A}
\sum_{i \in \mathcal{A}} q_i.
\end{equation}

We also examined whether accepted pseudo-labels were concentrated in particular GCD sky-type groups. For each true sky-type group $c$, true-group coverage was defined as the fraction of unlabelled samples from that group that were accepted:
\begin{equation}
\mathrm{Coverage}_c
=
\frac{
|\{i \in \mathcal{A}: y_i=c\}|
}{
|\{i \in \mathcal{U}: y_i=c\}|
}.
\end{equation}
To account for differences in sky-type-group prevalence within the unlabelled pool, we computed an accepted-group enrichment ratio,
\begin{equation}
\mathrm{Enrichment}_c
=
\frac{
|\{i \in \mathcal{A}: y_i=c\}|/A
}{
|\{i \in \mathcal{U}: y_i=c\}|/U
}.
\end{equation}
A value above one indicates that sky-type group $c$ was over-represented among accepted pseudo-labels relative to its prevalence in the unlabelled pool, while a value below one indicates under-representation.

Finally, for each predicted pseudo-label class $c$, we computed the precision of accepted pseudo-labels assigned to that class:
\begin{equation}
\mathrm{Precision}^{\mathrm{pseudo}}_c
=
\frac{
|\{i \in \mathcal{A}: \widehat{y}_i=c,\; y_i=c\}|
}{
|\{i \in \mathcal{A}: \widehat{y}_i=c\}|
}.
\end{equation}

Class-wise accepted counts were averaged over all five random seeds, including seeds in which zero samples from a class were accepted. Coverage and enrichment were computed from pooled counts across seeds, while predicted-class precision was computed over accepted pseudo-labels assigned to each class. If no accepted pseudo-labels were assigned to a class, the corresponding predicted-class precision was treated as undefined.

\section{Results}\label{sec:results}

\subsection{Supervised label-efficiency curve}\label{sec:results_supervised}

\begin{figure}[htbp]
    \centering
    \includegraphics[width=\linewidth]{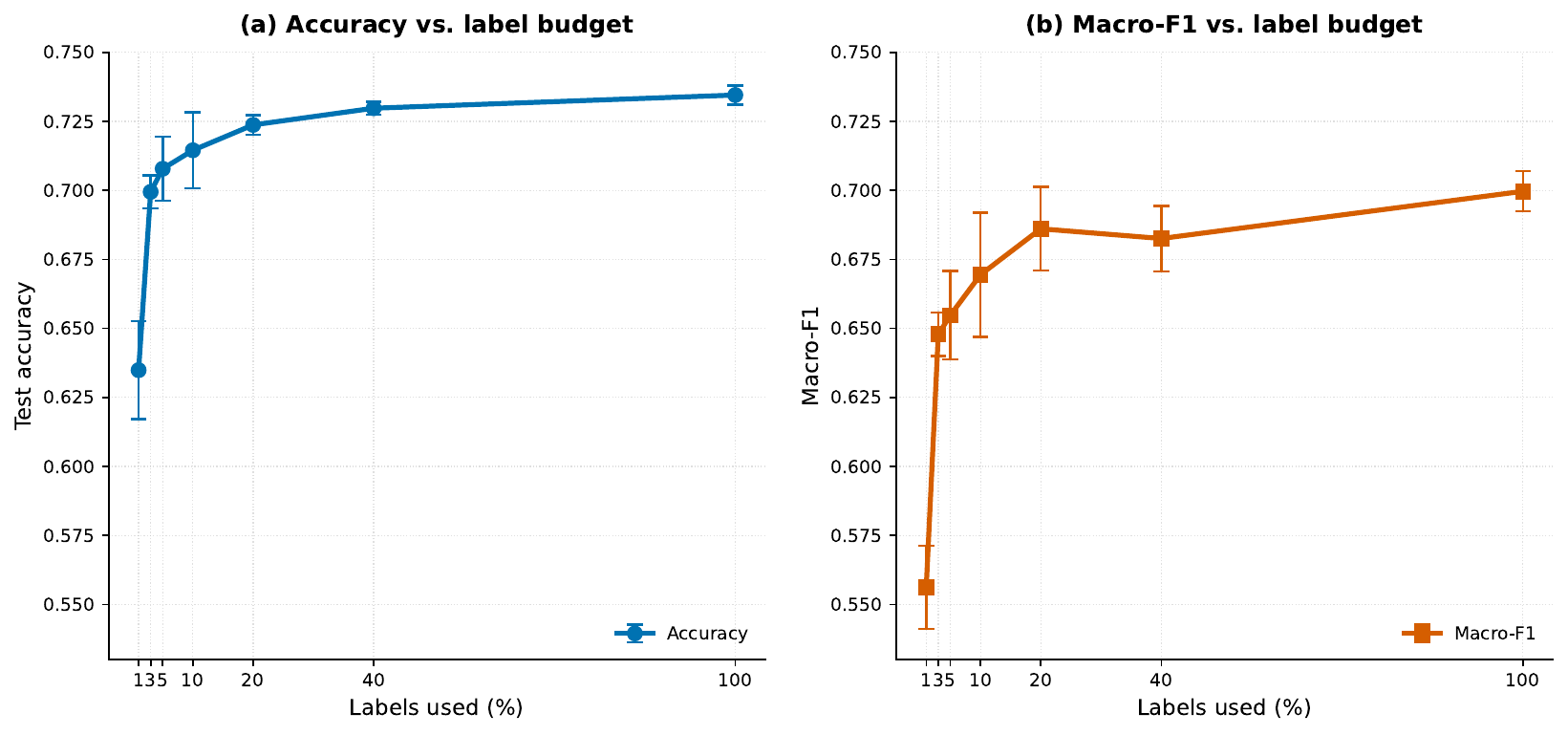}
    \caption{\textbf{Supervised label-budget performance of the ImageNet-pretrained ResNet50 baseline on GCD.} Panel (a) shows test accuracy and panel (b) shows macro-F1. Points denote the mean over five random seeds and error bars denote one standard deviation. Accuracy increases rapidly at low label budgets and begins to plateau from approximately $20$--$40\%$ labels, while macro-F1 shows a more gradual improvement with some seed-level variability.}
    \label{fig:supervised_label_budget_performance}
\end{figure}

We first evaluated the supervised transfer-learning baseline across label budgets to quantify how classification performance changes with decreasing annotation availability. Figure~\ref{fig:supervised_label_budget_performance} shows label-budget curves for test accuracy and macro-F1, while Table~\ref{tab:supervised_label_budget_metrics} reports accuracy, balanced accuracy, macro-F1, and weighted-F1 values.

Test accuracy increased rapidly in the low-label regime and then improved more gradually as additional labels were added. With $1\%$ of the training labels, corresponding to 100 labelled images, the model achieved $0.6349 \pm 0.0178$ accuracy, increasing to $0.7078 \pm 0.0117$ at $5\%$ labels, $0.7237 \pm 0.0035$ at $20\%$, and $0.7298 \pm 0.0024$ at $40\%$. Under full supervision, accuracy was $0.7346 \pm 0.0034$. Thus, $20\%$ of the available labels recovered most of the full-label accuracy, within approximately 1.1 percentage points of the $100\%$-label baseline, while $40\%$ labels reduced the gap to approximately 0.5 percentage points.

The class-balanced metrics showed a more gradual improvement and were not strictly monotonic across label budgets. Macro-F1 increased from $0.5562 \pm 0.0151$ at $1\%$ labels to $0.6548 \pm 0.0160$ at $5\%$, $0.6861 \pm 0.0152$ at $20\%$, and $0.6997 \pm 0.0073$ under full supervision. Balanced accuracy similarly increased from $0.5428 \pm 0.0199$ at $1\%$ labels to $0.6762 \pm 0.0075$ under full supervision. The small decrease in macro-F1 and balanced accuracy between $20\%$ and $40\%$ labels falls within the observed seed-level variability and should not be interpreted as a systematic degradation. Overall, the class-balanced metrics indicate that additional labelled data remained useful for improving performance across the sky-type groups, even after overall accuracy begins to show diminishing gains.

The gap between overall accuracy and class-balanced metrics suggests that label efficiency was not uniform across sky-type groups. Overall accuracy approached the full-label baseline relatively early, whereas macro-F1 and balanced accuracy improved more gradually. This indicates that the benefits of additional labels were not uniform across groups. The corresponding group-level behaviour is examined in Section~\ref{sec:results_per_class}.

\begin{table}[htbp]
\centering
\caption{\textbf{Supervised ResNet50 performance across labelled-data budgets.} Results are reported as mean $\pm$ standard deviation over five random seeds.}
\label{tab:supervised_label_budget_metrics}

\begin{tabular}{@{}rrrrrr@{}}
\toprule
\multicolumn{1}{c}{\shortstack{Labels\\(\%)}} &
\multicolumn{1}{c}{\shortstack{Labelled\\images}} &
\multicolumn{1}{c}{Accuracy} &
\multicolumn{1}{c}{\shortstack{Balanced\\accuracy}} &
\multicolumn{1}{c}{Macro-F1} &
\multicolumn{1}{c}{Weighted-F1} \\
\midrule
1 & 100 & $0.6349 \pm 0.0178$ & $0.5428 \pm 0.0199$ & $0.5562 \pm 0.0151$ & $0.6116 \pm 0.0201$ \\
3 & 299 & $0.6995 \pm 0.0060$ & $0.6291 \pm 0.0066$ & $0.6478 \pm 0.0078$ & $0.6888 \pm 0.0058$ \\
5 & 500 & $0.7078 \pm 0.0117$ & $0.6364 \pm 0.0137$ & $0.6548 \pm 0.0160$ & $0.6966 \pm 0.0128$ \\
10 & 1,000 & $0.7145 \pm 0.0138$ & $0.6517 \pm 0.0229$ & $0.6695 \pm 0.0226$ & $0.7054 \pm 0.0158$ \\
20 & 2,001 & $0.7237 \pm 0.0035$ & $0.6656 \pm 0.0156$ & $0.6861 \pm 0.0152$ & $0.7173 \pm 0.0083$ \\
40 & 3,999 & $0.7298 \pm 0.0024$ & $0.6598 \pm 0.0076$ & $0.6826 \pm 0.0119$ & $0.7184 \pm 0.0059$ \\
100 & 10,000 & $0.7346 \pm 0.0034$ & $0.6762 \pm 0.0075$ & $0.6997 \pm 0.0073$ & $0.7270 \pm 0.0075$ \\
\bottomrule
\end{tabular}
\end{table} 

\subsection{Comparison of supervised, active-learning, and pseudo-labeling strategies}
\label{sec:results_method_comparison}

\begin{figure}[htbp]
    \centering
    \includegraphics[width=\linewidth]{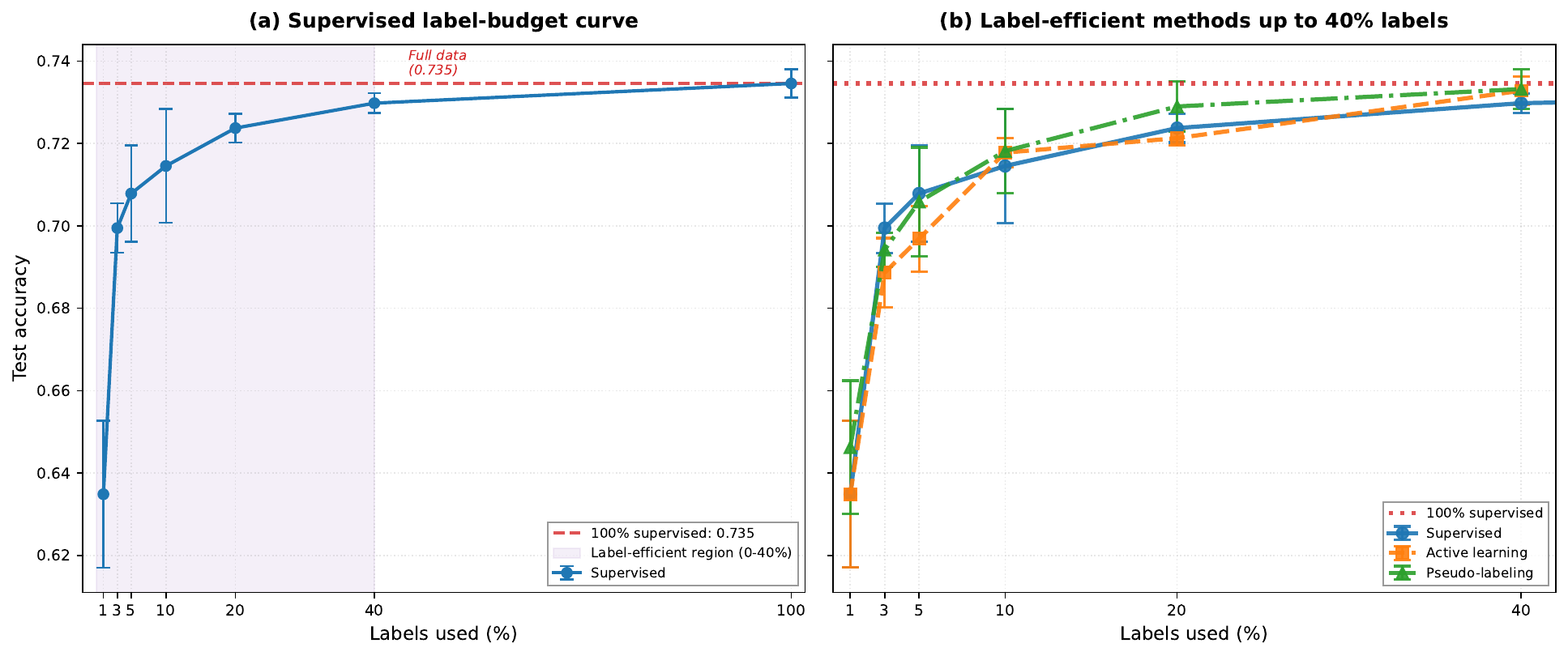} 
    \caption{\textbf{Label-efficiency results on GCD.} Points show mean test accuracy over five random seeds and error bars show one standard deviation. The horizontal reference line indicates the $100\%$-label supervised result. (a) Supervised ResNet50 performance across labelled-data budgets from $1\%$ to $100\%$. (b) Comparison of supervised stratified random sampling, uncertainty-based active learning, and high-confidence pseudo-labeling up to $40\%$ labels.}
    \label{fig:method_comparison_accuracy}
\end{figure}

We next compared the supervised stratified random-sampling, uncertainty-based active learning, and high-confidence pseudo-labeling under matched label budgets. Figure~\ref{fig:method_comparison_accuracy} shows the test-accuracy curves up to $40\%$ labels, and Table~\ref{tab:method_comparison_metrics} reports the corresponding accuracy, macro-F1, balanced accuracy, and weighted-F1 values.

\begin{table}[htbp]
\centering
\caption{\textbf{Comparison of supervised learning, active learning, and pseudo-labeling across labelled-data budgets.} Results are reported as mean $\pm$ standard deviation over five random seeds.}
\label{tab:method_comparison_metrics}
\small
\setlength{\tabcolsep}{4.5pt}
\renewcommand{\arraystretch}{1.1}
\begin{tabular}{@{}lccccc@{}}
\toprule
\multicolumn{1}{c}{Methods} &
\multicolumn{1}{c}{\shortstack{Labels\\(\%)}} &
\multicolumn{1}{c}{Accuracy} &
\multicolumn{1}{c}{Macro-F1} &
\multicolumn{1}{c}{\shortstack{Balanced\\accuracy}} &
\multicolumn{1}{c}{Weighted-F1} \\
\midrule
Supervised & 1 & $0.635 \pm 0.018$ & $0.556 \pm 0.015$ & $0.543 \pm 0.020$ & $0.612 \pm 0.020$ \\
Active learning & 1 & $0.635 \pm 0.018$ & $0.556 \pm 0.015$ & $0.543 \pm 0.020$ & $0.612 \pm 0.020$ \\
Pseudo-labeling & 1 & $0.646 \pm 0.016$ & $0.573 \pm 0.016$ & $0.563 \pm 0.019$ & $0.623 \pm 0.017$ \\
\midrule
Supervised & 3 & $0.699 \pm 0.006$ & $0.648 \pm 0.008$ & $0.629 \pm 0.007$ & $0.689 \pm 0.006$ \\
Active learning & 3 & $0.689 \pm 0.008$ & $0.651 \pm 0.008$ & $0.625 \pm 0.016$ & $0.683 \pm 0.009$ \\
Pseudo-labeling & 3 & $0.694 \pm 0.004$ & $0.629 \pm 0.010$ & $0.617 \pm 0.005$ & $0.675 \pm 0.007$ \\
\midrule
Supervised & 5 & $0.708 \pm 0.012$ & $0.655 \pm 0.016$ & $0.636 \pm 0.014$ & $0.697 \pm 0.013$ \\
Active learning & 5 & $0.697 \pm 0.008$ & $0.666 \pm 0.010$ & $0.645 \pm 0.010$ & $0.692 \pm 0.007$ \\
Pseudo-labeling & 5 & $0.706 \pm 0.013$ & $0.654 \pm 0.013$ & $0.638 \pm 0.011$ & $0.695 \pm 0.011$ \\
\midrule
Supervised & 10 & $0.715 \pm 0.014$ & $0.669 \pm 0.023$ & $0.652 \pm 0.023$ & $0.705 \pm 0.016$ \\
Active learning & 10 & $0.718 \pm 0.003$ & $0.689 \pm 0.009$ & $0.664 \pm 0.007$ & $0.713 \pm 0.004$ \\
Pseudo-labeling & 10 & $0.718 \pm 0.010$ & $0.675 \pm 0.013$ & $0.659 \pm 0.011$ & $0.710 \pm 0.010$ \\
\midrule
Supervised & 20 & $0.724 \pm 0.004$ & $0.686 \pm 0.015$ & $0.666 \pm 0.016$ & $0.717 \pm 0.008$ \\
Active learning & 20 & $0.721 \pm 0.002$ & $0.685 \pm 0.004$ & $0.663 \pm 0.005$ & $0.714 \pm 0.002$ \\
Pseudo-labeling & 20 & $0.729 \pm 0.006$ & $0.691 \pm 0.011$ & $0.670 \pm 0.007$ & $0.722 \pm 0.007$ \\
\midrule
Supervised & 40 & $0.730 \pm 0.002$ & $0.683 \pm 0.012$ & $0.660 \pm 0.008$ & $0.718 \pm 0.006$ \\
Active learning & 40 & $0.733 \pm 0.003$ & $0.697 \pm 0.007$ & $0.676 \pm 0.009$ & $0.726 \pm 0.005$ \\
Pseudo-labeling & 40 & $0.733 \pm 0.005$ & $0.695 \pm 0.012$ & $0.672 \pm 0.013$ & $0.725 \pm 0.008$ \\
\midrule
Supervised & 100 & $0.735 \pm 0.003$ & $0.700 \pm 0.007$ & $0.676 \pm 0.007$ & $0.727 \pm 0.007$ \\
\bottomrule
\end{tabular}
\end{table}

Overall, the three strategies produced broadly similar test-accuracy curves, indicating that the frozen ResNet50 transfer-learning baseline was already highly label-efficient on GCD. With only $1\%$ of the training labels, supervised learning achieved $0.635 \pm 0.018$ accuracy, while pseudo-labeling increased this to $0.646 \pm 0.016$. Active learning was identical to the supervised baseline at $1\%$ because both methods used the same initial labelled subset before any acquisition step.

At $3\%$ and $5\%$ labels, supervised learning achieved the highest mean accuracy, with $0.699 \pm 0.006$ and $0.708 \pm 0.012$, respectively. Active learning had lower accuracy at these budgets, although its macro-F1 and balanced accuracy were competitive, particularly at $5\%$ labels. From $10\%$ labels onward, the methods became increasingly close. At $10\%$ labels, active learning and pseudo-labeling both reached approximately $0.718$ accuracy, compared with $0.715 \pm 0.014$ for the supervised baseline. At $40\%$ labels, active learning and pseudo-labeling both achieved $0.733$ mean accuracy, close to the $100\%$-label supervised reference of $0.735 \pm 0.003$.

Class-balanced metrics showed small advantages for the label-efficient methods at some budgets. Active learning had the highest macro-F1 at $3\%$, $5\%$, $10\%$, and $40\%$ labels, while pseudo-labeling had the highest macro-F1 at $1\%$ and $20\%$ labels. However, these differences were modest in absolute terms. For example, at $40\%$ labels, active learning improved macro-F1 from $0.683 \pm 0.012$ to $0.697 \pm 0.007$ relative to supervised sampling, while test accuracy increased only from $0.730 \pm 0.002$ to $0.733 \pm 0.003$.

To control for the additional optimisation introduced by the second pseudo-labeling stage, we also trained a continued-supervised model from the same first-stage checkpoint for the same number of optimiser updates. As shown in Appendix~\ref{app:pseudo_control}, pseudo-labeling did not consistently outperform this update-matched control, with semi-supervised minus control differences remaining close to zero across label budgets. Thus, the apparent gains of pseudo-labeling over the first-stage supervised model should be interpreted cautiously.

These results show that simple uncertainty sampling and high-confidence pseudo-labeling remained competitive with supervised stratified sampling and produced small gains for some metrics and label budgets. However, neither method produced a large or consistent improvement over the strong transfer-learning baseline. The diagnostic analyses in Sections~\ref{sec:results_pseudo_quality} and~\ref{sec:results_active_queries} examine possible reasons for these limited gains.

\subsection{Per-class performance and confusion patterns} \label{sec:results_per_class}

To examine which sky-type groups contributed most to the remaining performance differences, we analysed per-class F1 scores. Table~\ref{tab:per_class_f1_selected_methods} reports results for the three learning strategies at 10\% and 40\% labels. These budgets were selected to represent an intermediate low-label setting and the largest label-efficient budget considered before full supervision.

\begin{table}[htbp]
\centering
\caption{\textbf{Per-class F1 scores for supervised learning, active learning, and pseudo-labeling at selected label budgets.} Results are reported as mean $\pm$ standard deviation over five random seeds.}
\label{tab:per_class_f1_selected_methods}

\footnotesize
\setlength{\tabcolsep}{3pt}

\begin{tabular}{@{}lcccccc@{}}
\toprule
& \multicolumn{2}{c}{Supervised}
& \multicolumn{2}{c}{Active learning}
& \multicolumn{2}{c}{Pseudo-labeling} \\
\cmidrule(lr){2-3}
\cmidrule(lr){4-5}
\cmidrule(l){6-7}
Class
& 10\% & 40\%
& 10\% & 40\%
& 10\% & 40\% \\
\midrule
Altocumulus
& $0.624 \pm 0.028$ & $0.625 \pm 0.020$
& $0.611 \pm 0.019$ & $0.586 \pm 0.012$
& $0.649 \pm 0.020$ & $0.614 \pm 0.023$ \\
Cirrus
& $0.663 \pm 0.023$ & $0.716 \pm 0.013$
& $0.702 \pm 0.015$ & $0.711 \pm 0.012$
& $0.663 \pm 0.021$ & $0.712 \pm 0.017$ \\
Clear sky
& $0.957 \pm 0.012$ & $0.974 \pm 0.004$
& $0.964 \pm 0.007$ & $0.972 \pm 0.004$
& $0.957 \pm 0.007$ & $0.972 \pm 0.003$ \\
Cumulonimbus
& $0.738 \pm 0.017$ & $0.744 \pm 0.015$
& $0.711 \pm 0.015$ & $0.751 \pm 0.009$
& $0.741 \pm 0.015$ & $0.749 \pm 0.009$ \\
Cumulus
& $0.732 \pm 0.022$ & $0.746 \pm 0.009$
& $0.730 \pm 0.014$ & $0.750 \pm 0.013$
& $0.730 \pm 0.017$ & $0.748 \pm 0.008$ \\
Mixed
& $0.394 \pm 0.086$ & $0.376 \pm 0.066$
& $0.497 \pm 0.033$ & $0.511 \pm 0.046$
& $0.399 \pm 0.075$ & $0.471 \pm 0.060$ \\
Stratocumulus
& $0.578 \pm 0.025$ & $0.597 \pm 0.017$
& $0.606 \pm 0.011$ & $0.598 \pm 0.020$
& $0.584 \pm 0.016$ & $0.597 \pm 0.022$ \\
\bottomrule
\end{tabular}

\vspace{1.5pt}
\footnotesize
\emph{Note:} Shortened labels correspond to the GCD sky-type groups defined in Section~\ref{sec:dataset}.
\end{table}

Clear sky was consistently the easiest group, with F1 scores close to $0.96$--$0.97$ across methods and label budgets. Cumulus and Cumulonimbus also achieved relatively strong performance, with F1 scores around $0.71$--$0.75$. These groups contributed substantially to the high overall accuracy observed in the label-budget curves.

The most difficult groups were Mixed, Stratocumulus, and Altocumulus. Mixed had the lowest F1 scores across the selected budgets, reflecting the visual heterogeneity of this group. Active learning improved Mixed substantially relative to supervised sampling, increasing the F1 score from $0.394 \pm 0.086$ to $0.497 \pm 0.033$ at $10\%$ labels and from $0.376 \pm 0.066$ to $0.511 \pm 0.046$ at $40\%$ labels. Pseudo-labeling also improved Mixed at $40\%$ labels, reaching $0.471 \pm 0.060$, although the gain was smaller than that of active learning.

Altocumulus remained challenging across methods, with F1 scores around $0.59$--$0.65$ at the selected budgets. Stratocumulus also remained difficult, with F1 scores around $0.58$--$0.61$. These results indicate that the gap between overall accuracy and macro-F1 was driven by weaker performance on heterogeneous or visually overlapping groups rather than by uniform errors across all categories.

The class-level results are consistent with the diagnostic analyses in Sections~\ref{sec:results_pseudo_quality} and~\ref{sec:results_active_queries}. High-confidence pseudo-labeling tended to select easier examples, while active learning queried uncertain samples from difficult groups, including Mixed and the Stratocumulus--Cumulonimbus boundary. However, these targeted acquisitions produced only modest improvements, suggesting that the remaining errors involve ambiguous or visually overlapping cases. Row-normalised confusion matrices, reported in Appendix~\ref{app:confusion_matrices}, further illustrate persistent confusion involving Mixed scenes and confusion between Stratocumulus and Cumulonimbus groups.

\subsection{Pseudo-label quality and selection bias}
\label{sec:results_pseudo_quality}

\begin{figure}[htbp]
\centering
\includegraphics[width=\linewidth]{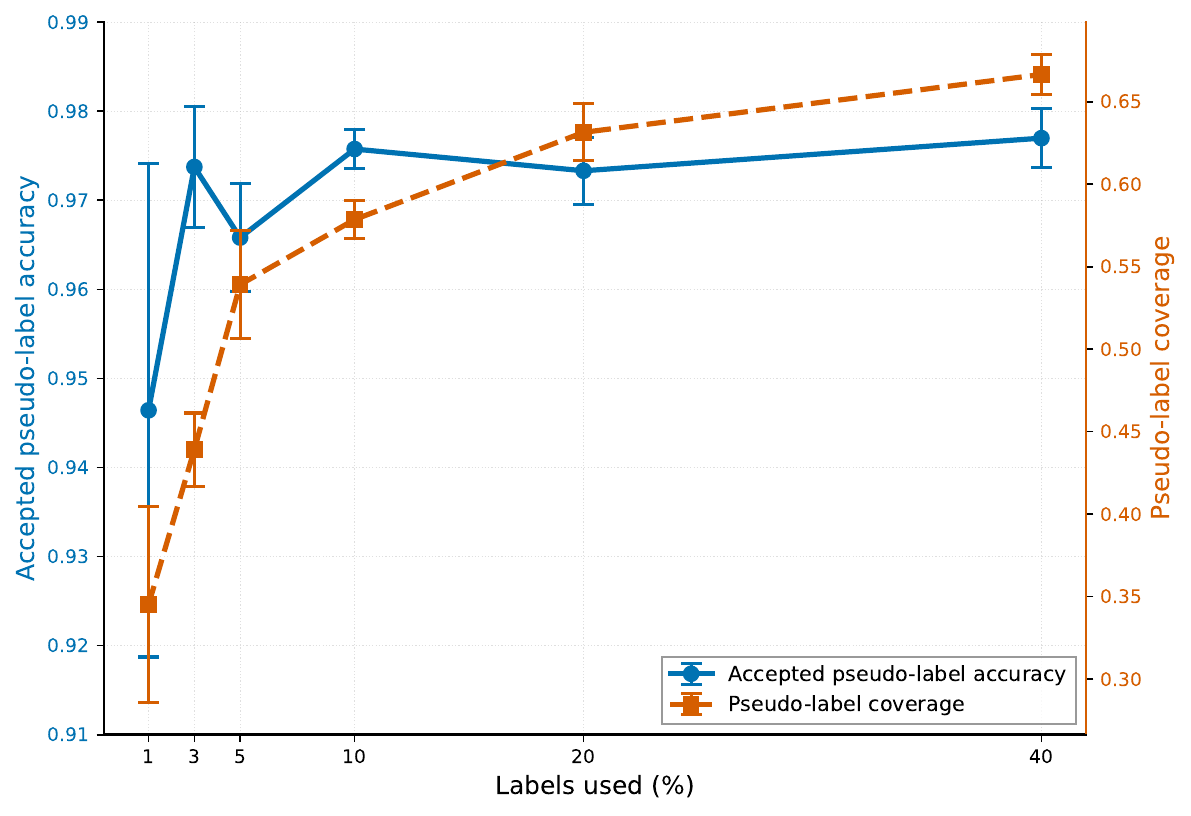}
\caption{\textbf{Pseudo-label quality on the unlabelled training pool.} Accepted pseudo-label accuracy denotes the fraction of retained pseudo-labels that matched the ground-truth labels, using training labels only for post hoc diagnostic evaluation. Coverage denotes the fraction of the unlabelled pool assigned a pseudo-label with confidence at least $\tau=0.95$. Points show the mean over five random seeds and error bars show one standard deviation.}
\label{fig:pseudo_label_quality}
\end{figure}

The pseudo-labeling results in Section~\ref{sec:results_method_comparison} showed limited improvements over the supervised and continued-supervised baselines. To understand why, we analysed the pseudo-labels accepted using the confidence threshold $\tau=0.95$. Figure~\ref{fig:pseudo_label_quality} shows accepted pseudo-label accuracy and coverage as a function of label budget, while Table~\ref{tab:pseudo_label_quality} reports the corresponding numerical values. Class-relative pseudo-label selection diagnostics are reported in Appendix~\ref{app:pseudo_label_distribution}.

\begin{table}[htbp]
\centering
\caption{\textbf{Pseudo-label quality across labelled-data budgets at threshold $\tau=0.95$.} Results are reported as mean $\pm$ standard deviation over five random seeds. Coverage is the fraction of the unlabelled pool assigned a pseudo-label. Accepted accuracy is the accuracy of predictions retained as pseudo-labels, while rejected accuracy is the accuracy of predictions below the confidence threshold, computed only for diagnostic purposes.}
\label{tab:pseudo_label_quality}
\scriptsize
\setlength{\tabcolsep}{3.5pt}
\renewcommand{\arraystretch}{1.15}
\begin{tabular}{@{}rrrrrrr@{}}
\toprule
\multicolumn{1}{c}{\shortstack{Labels\\(\%)}} &
\multicolumn{1}{c}{\shortstack{Labelled\\images}} &
\multicolumn{1}{c}{\shortstack{Pseudo\\labels}} &
\multicolumn{1}{c}{Coverage} &
\multicolumn{1}{c}{\shortstack{Accepted\\accuracy}} &
\multicolumn{1}{c}{\shortstack{Rejected\\accuracy}} &
\multicolumn{1}{c}{\shortstack{Accepted\\confidence}} \\
\midrule
1 & 100 & $3418 \pm 587$ & $0.345 \pm 0.059$ & $0.946 \pm 0.028$ & $0.664 \pm 0.012$ & $0.984 \pm 0.001$ \\
3 & 299 & $4260 \pm 215$ & $0.439 \pm 0.022$ & $0.974 \pm 0.007$ & $0.730 \pm 0.011$ & $0.986 \pm 0.001$ \\
5 & 500 & $5122 \pm 311$ & $0.539 \pm 0.033$ & $0.966 \pm 0.006$ & $0.692 \pm 0.019$ & $0.989 \pm {<}0.001$ \\
10 & 1000 & $5207 \pm 105$ & $0.579 \pm 0.012$ & $0.976 \pm 0.002$ & $0.719 \pm 0.013$ & $0.990 \pm 0.001$ \\
20 & 2001 & $5051 \pm 139$ & $0.631 \pm 0.017$ & $0.973 \pm 0.004$ & $0.721 \pm 0.010$ & $0.991 \pm 0.001$ \\
40 & 3999 & $4000 \pm 73$ & $0.666 \pm 0.012$ & $0.977 \pm 0.003$ & $0.716 \pm 0.011$ & $0.992 \pm 0.001$ \\
\bottomrule
\end{tabular}
\end{table}

The accepted pseudo-labels were substantially more reliable than the rejected predictions. Accepted pseudo-label accuracy ranged from $0.946 \pm 0.028$ at 1\% labels to $0.977 \pm 0.003$ at 40\% labels, whereas the accuracy of rejected predictions remained much lower. Coverage increased from $0.345 \pm 0.059$ to $0.666 \pm 0.012$ as the labelled-data budget increased. The absolute number of accepted pseudo-labels at 20\% and 40\% labels was nevertheless lower than at 10\% labels, despite higher coverage, because the unlabelled pool was smaller at the larger labelled-data budgets.

However, high pseudo-label quality did not translate into consistent gains over the update-matched continued-supervised control. This indicates that the limitation was not only pseudo-label noise, but also which samples were selected. The class-relative diagnostics in Appendix~\ref{app:pseudo_label_distribution} show that Clear sky was consistently over-represented among accepted pseudo-labels, with enrichment above one at every budget, while Mixed was consistently under-represented, with low true-group coverage and enrichment below one. This supports the conclusion that high-confidence pseudo-labeling selected easy, visually distinctive examples more readily than ambiguous mixed-cloud scenes.

This selection pattern helps explain why high-confidence pseudo-labeling did not consistently improve macro-F1 or balanced accuracy. The pseudo-labels were highly accurate, but they mainly added examples that the model already classified confidently. Consequently, they contributed relatively few additional examples from the groups associated with the remaining classification errors, particularly Mixed and visually similar groups such as Stratocumulus and Cumulonimbus. Thus, in this setting, the main limitation of pseudo-labeling was not simply pseudo-label noise, but the limited additional information provided by confidence-selected pseudo-labels.

\subsection{Active-learning query behaviour}
\label{sec:results_active_queries}

\begin{figure}[htbp]
    \centering
    \includegraphics[width=\linewidth]{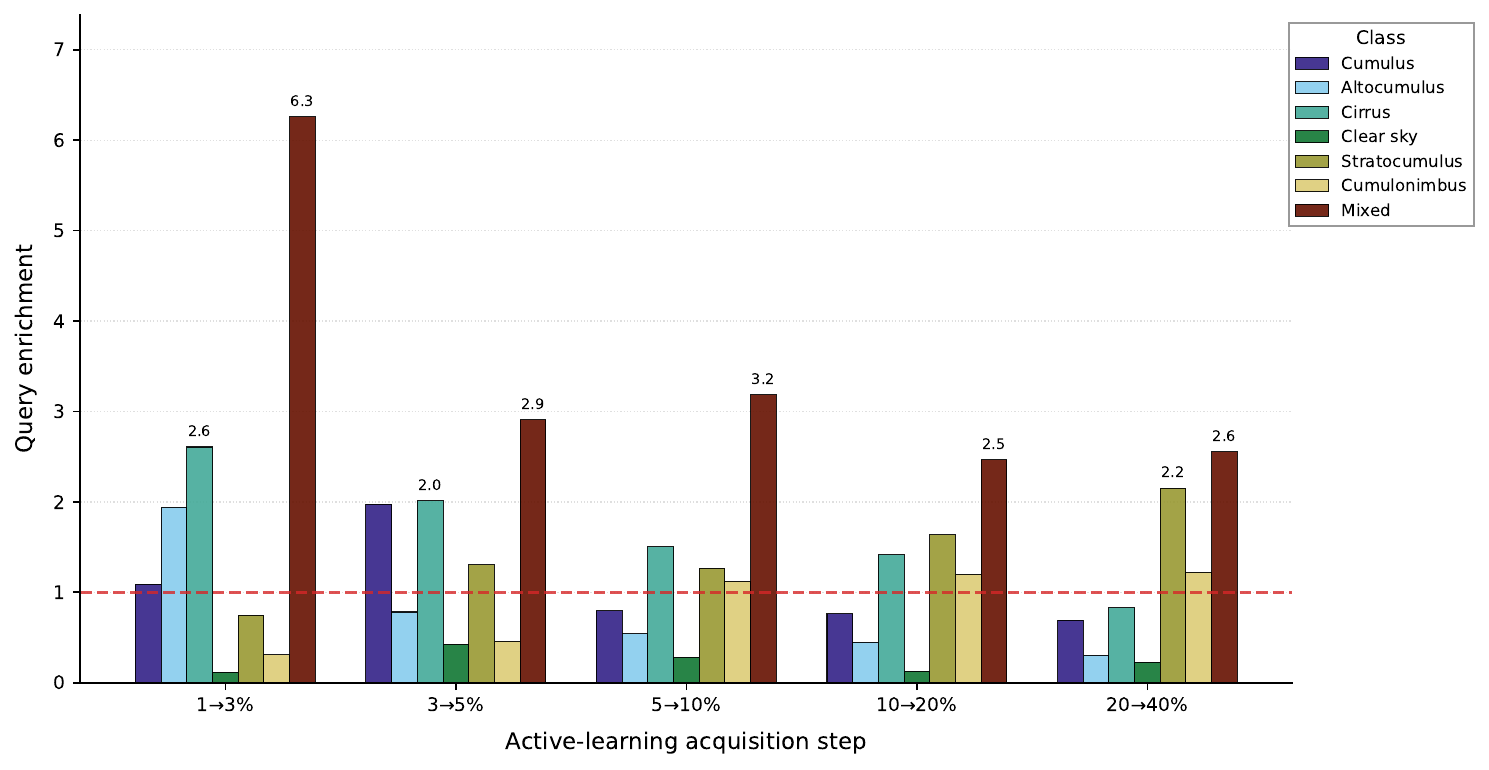}
    \caption{\textbf{Sky-type group enrichment among samples queried by uncertainty-based active learning.} Enrichment is computed relative to the remaining unlabelled pool immediately before each acquisition step and averaged over five random seeds. Values above one indicate that a sky-type group was over-represented among queried samples relative to its availability in the current unlabelled pool, while values below one indicate under-representation. The dashed horizontal line indicates an enrichment ratio of one.Shortened labels correspond to the GCD sky-type groups defined in Section~\ref{sec:dataset}.}
    \label{fig:active_query_enrichment}
\end{figure}

We analysed the samples selected by uncertainty-based active learning to characterise the acquisition behaviour of the method. Figure~\ref{fig:active_query_enrichment} shows the active-query enrichment ratio for each sky-type group and acquisition step. Enrichment was computed relative to the remaining unlabelled pool immediately before each acquisition step, rather than relative to the original full training distribution. A value of one therefore indicates that the queried fraction for a class matched its availability in the current unlabelled pool, while values above one indicate over-representation and values below one indicate under-representation. The corresponding queried counts, queried fractions, pre-acquisition pool fractions, and enrichment ratios are reported in Appendix~\ref{app:active_query_distribution}.

Uncertainty sampling did not simply reproduce the distribution of the remaining unlabelled pool. At the earliest acquisition step, from $1\%$ to $3\%$ labels, Mixed was strongly enriched, with an enrichment ratio of $6.26 \pm 1.23$, despite representing only a small fraction of the available unlabelled pool. Cirrus and Altocumulus were also over-represented at this stage, with enrichment ratios of $2.61 \pm 0.80$ and $1.94 \pm 0.74$, respectively. In contrast, Clear sky was consistently under-represented across acquisition steps, indicating that the model rarely regarded Clear sky images as highly uncertain.

At later acquisition steps, the queried samples became increasingly concentrated in the visually confusable Stratocumulus and Cumulonimbus groups. During the $20\%\to40\%$ acquisition step, Stratocumulus and Cumulonimbus accounted for the largest queried fractions, $0.360 \pm 0.020$ and $0.365 \pm 0.018$, respectively. Their corresponding enrichment ratios were $2.15 \pm 0.10$ and $1.22 \pm 0.07$. Mixed also remained enriched at this stage, with an enrichment ratio of $2.55 \pm 0.21$, although its queried fraction was smaller because few Mixed samples remained in the unlabelled pool. 

The analysis shows that maximum-softmax uncertainty sampling targeted sky-type groups associated with visual ambiguity, particularly Mixed in the early acquisition steps and the Stratocumulus--Cumulonimbus boundary at higher budgets. This pattern is consistent with the confusion-matrix and per-class analyses, where Mixed scenes and the Stratocumulus and Cumulonimbus groups remained among the more difficult cases. Despite this targeted acquisition behaviour, active learning produced only limited improvement over the supervised label-budget baseline. This suggests that the remaining errors were not resolved simply by adding uncertain examples, possibly because the queried samples were difficult boundary cases, redundant within feature space, or insufficient to overcome the strong baseline provided by ImageNet-pretrained transfer learning. More sophisticated acquisition strategies, such as uncertainty combined with diversity, class-balanced uncertainty sampling, or methods explicitly targeting persistent confusion pairs, may be needed to obtain larger gains.

\section{Discussion}\label{sec:discussion}

In this study, we investigated label-efficient ground-based cloud classification on GCD using a frozen ImageNet-pretrained ResNet50 backbone. We compared supervised stratified sampling, uncertainty-based active learning, and high-confidence pseudo-labeling under matched label budgets and a fixed training protocol. The main finding is that the supervised transfer-learning baseline was already highly label-efficient. Test accuracy increased rapidly from $1\%$ to $5\%$ labels and then improved more gradually, with diminishing gains at larger label budgets. With $20$--$40\%$ of the training labels, performance approached that obtained with the full labelled set. This suggests that generic visual features learned from ImageNet can transfer effectively to ground-based cloud imagery, even when the labelled pool is severely limited.

However, class-balanced metrics showed a more nuanced picture. Although overall accuracy approached the full-label result relatively early, macro-F1 and balanced accuracy improved more gradually as additional labels were added. This indicates that the benefits of additional annotation were not uniform across sky-type groups. The per-class analysis confirmed this interpretation. Clear sky was classified accurately across all budgets, whereas Mixed, Altocumulus, and Stratocumulus remained persistently challenging. Mixed scenes were particularly difficult, likely because they contain visual characteristics associated with multiple sky-type groups. The persistent confusion between the Stratocumulus/Stratus/Altostratus and Cumulonimbus/Nimbostratus groups further suggests that some errors arise from substantial visual overlap rather than simply insufficient labelled data.

The active-learning results show that maximum-softmax uncertainty sampling was competitive with supervised stratified sampling and provided small improvements for some class-balanced metrics. In particular, active learning improved performance on Mixed at selected budgets and achieved the highest macro-F1 at several label budgets. However, these gains were modest and did not translate into a large improvement in overall accuracy. The query-enrichment analysis showed that uncertainty sampling did not simply reproduce the distribution of the remaining unlabelled pool. Instead, it strongly enriched Mixed during the earliest acquisition step and later concentrated on the visually confusable Stratocumulus and Cumulonimbus groups, while consistently under-selecting Clear sky. Thus, the limited gains were not because active learning failed to query difficult sky-type groups. Rather, the queried samples may have represented difficult boundary cases, partially redundant examples, or insufficient additional information to overcome the strong transfer-learning baseline.

High-confidence pseudo-labeling also produced only limited gains. At higher budgets, it slightly improved accuracy or macro-F1 relative to the first-stage supervised model, but the controlled comparison showed that pseudo-labeling did not consistently outperform an update-matched continued-supervised control. This indicates that part of the apparent benefit may have resulted from additional optimisation rather than from the pseudo-labelled samples themselves. The pseudo-label quality analysis showed that accepted pseudo-labels were substantially more accurate than rejected predictions, so the limitation was not primarily pseudo-label noise. Instead, the class-relative diagnostics showed a selection effect: the confidence threshold preferentially accepted easier or more visually distinctive sky-type groups, especially Clear sky, while Mixed remained under-represented relative to its difficulty. As a result, pseudo-labeling added many reliable examples, but not necessarily the examples most likely to improve class-balanced performance.

Overall, these findings suggest that simple label-efficient strategies provide limited additional benefit when the supervised transfer-learning baseline is already strong. A relatively small labelled subset may be sufficient to obtain strong overall accuracy on GCD, but improving performance on ambiguous sky-type groups requires more targeted approaches. The results also show why diagnostic analyses are important: pseudo-labeling and active learning can produce similar aggregate performance while behaving differently in terms of sample selection. Pseudo-labeling mainly adds high-confidence examples, whereas uncertainty sampling actively queries ambiguous examples. Nevertheless, neither strategy fully resolved the persistent group-level errors.

Several directions could improve upon these baselines. For pseudo-labeling, class-aware thresholds, calibration-aware confidence scores, or methods that explicitly balance pseudo-label selection across sky-type groups may reduce the bias toward easy high-confidence examples. For active learning, uncertainty could be combined with diversity, representation-space coverage, or class-conditional acquisition objectives to reduce redundant selections and improve coverage of difficult regions. Hybrid active semi-supervised approaches may also be useful, for example by using active learning to label difficult regions of the feature space while using pseudo-labeling for reliable high-confidence samples.

This study has several limitations. First, the benchmark used a single frozen backbone architecture and a fixed preprocessing pipeline. This design enabled a controlled comparison of learning strategies, but other architectures, fine-tuning protocols, or augmentation strategies may yield different absolute performance. Second, the active-learning and pseudo-labeling methods were deliberately simple baselines; more advanced strategies may perform differently. Third, the analysis was restricted to GCD, so the conclusions should be validated on additional ground-based cloud datasets with different imaging conditions, class definitions, and geographic settings. Despite these limitations, the results provide a reproducible benchmark for label-efficient ground-based cloud classification and highlight the importance of class-level diagnostics when evaluating active and semi-supervised learning methods.

\section{Conclusion}\label{sec:conclusion}

This study presented a systematic benchmark of label-efficient learning strategies for ground-based cloud classification on GCD. Under matched label budgets and a fixed training protocol, supervised transfer learning with a frozen ImageNet-pretrained ResNet50 backbone proved highly label-efficient. Test accuracy increased rapidly at low label budgets, and using only $20$--$40\%$ of the available training labels approached the performance obtained with the full labelled set.

Active learning and high-confidence pseudo-labeling provided only limited additional benefit over the supervised baseline. The results show that the main challenge was not simply the number of labelled images, but the difficulty of improving performance on visually ambiguous sky-type groups. Pseudo-labeling produced reliable high-confidence labels, but these were biased toward easier sky-type groups and did not consistently outperform an update-matched continued-supervised control. In contrast, uncertainty-based active learning did query difficult groups, including Mixed and the Stratocumulus--Cumulonimbus boundary, but these targeted acquisitions produced only modest improvements in overall performance.

Overall, these results provide a reproducible benchmark for label-efficient ground-based cloud classification and show that strong transfer-learning baselines are difficult to improve with simple active or semi-supervised strategies. Future gains are likely to require more targeted methods, such as class-aware pseudo-labeling, uncertainty sampling combined with diversity, or hybrid active semi-supervised approaches designed specifically for visually ambiguous sky-type groups.

\section*{Declarations}

\bmhead{Funding}
E.B.D. is funded by the European Union's Horizon 2020 research and innovation programme under the Marie Sk{\l}odowska-Curie grant agreement n\textsuperscript{o} 101034255.

\bmhead{Competing interests}
The authors have no competing interests.

\bmhead{Ethics approval and consent to participate}
Not applicable.

\bmhead{Consent for publication}
Not applicable.

\bmhead{Data availability}
The Ground-based Cloud Dataset (GCD) used in this study is publicly available at
\url{https://github.com/shuangliutjnu/TJNU-Ground-based-Cloud-Dataset}.

\bmhead{Materials availability}
Not applicable.

\bmhead{Code availability}
The code used for the experiments is available at: \url{https://github.com/EstherBD/Label-Efficient-Ground-Based-Cloud-Classification-on-GCD.git}

\bmhead{Author contribution}
E.B.D. designed the study, implemented the experiments, analysed the results, and drafted the manuscript. V.B.D. contributed to the study design, interpretation of the results, and revision of the manuscript. B.Z. supervised the project, contributed to the interpretation of the results, and revised the manuscript.

\begin{appendices}

\section{Supervised confusion matrices}\label{app:confusion_matrices}

\begin{figure}[!t]
\centering
\includegraphics[width=\linewidth]{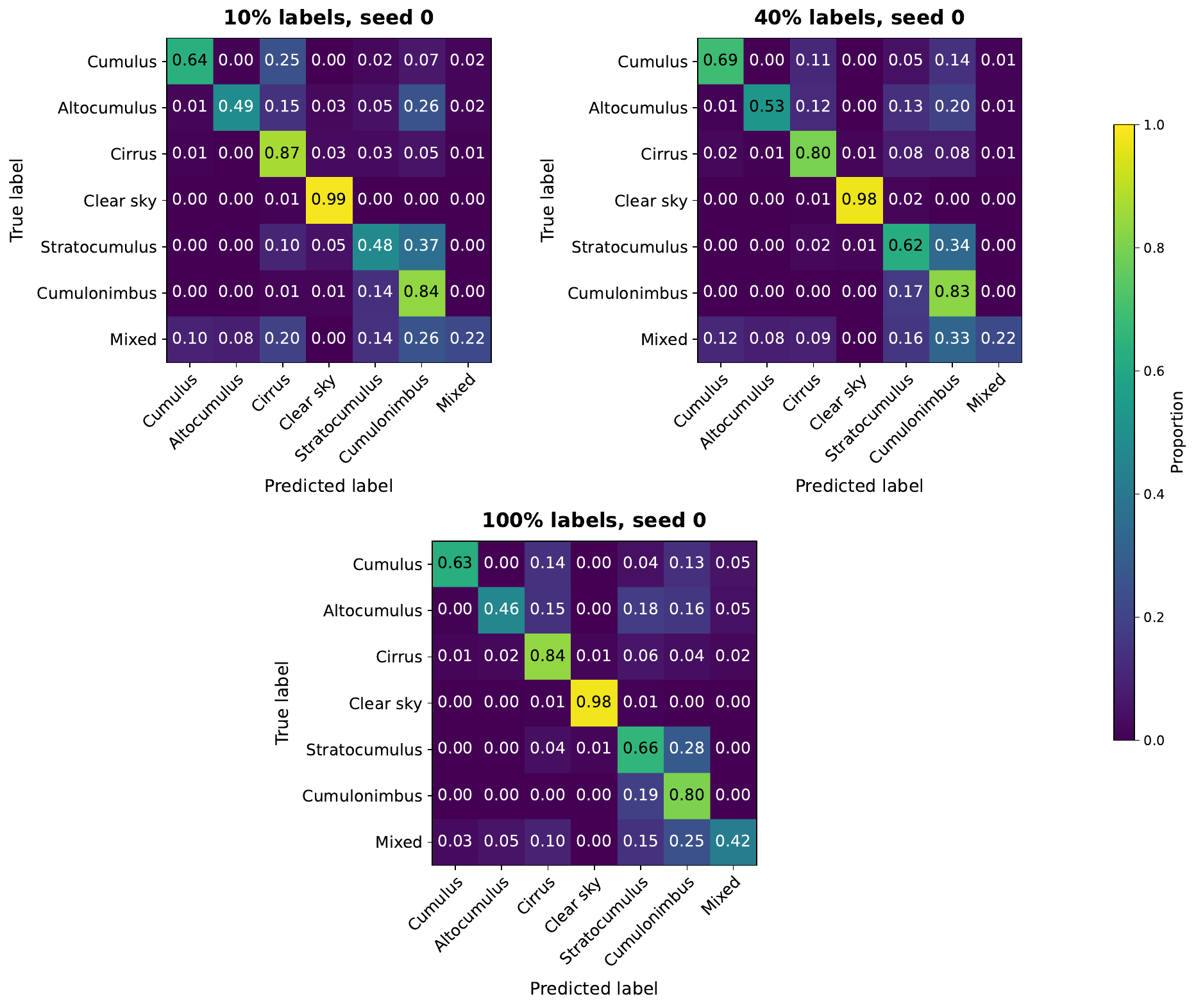}
\caption{\textbf{Row-normalised confusion matrices for the supervised ResNet50 baseline at selected label budgets for seed $0$.} Rows correspond to true labels and columns to predicted labels. Shortened class labels are used for readability and correspond to the GCD sky-type groups defined in Section~\ref{sec:dataset}. Clear sky is classified accurately across all selected budgets, whereas the main persistent errors involve Mixed scenes and confusion between Stratocumulus and Cumulonimbus groups.}
\label{fig:supervised_confusion_matrices}
\end{figure}

Figure~\ref{fig:supervised_confusion_matrices} reports row-normalised confusion matrices for the supervised ResNet50 baseline at selected label budgets. These matrices complement the per-class F1 analysis in Section~\ref{sec:results_per_class} by showing the main error patterns between GCD sky-type groups. Across the selected label budgets, Clear sky is classified with high recall, whereas the main persistent errors involve Mixed scenes and confusion between visually similar groups, particularly Stratocumulus/Stratus/Altostratus and Cumulonimbus/Nimbostratus.

\section{Controlled pseudo-labeling results}
\label{app:pseudo_control}

Table~\ref{tab:pseudo_control} reports the controlled pseudo-labeling results. The semi-supervised model was compared with both the first-stage labelled-only model and an update-matched continued-supervised control initialised from the same first-stage checkpoint. This control separates the effect of pseudo-label information from the effect of additional optimisation.

\begin{table}[htbp]
\centering
\caption{\textbf{Controlled pseudo-labeling results at threshold $\tau=0.95$.} Results are reported as mean $\pm$ standard deviation over five random seeds. The continued-supervised control was initialised from the same first-stage checkpoint as the semi-supervised model and trained for the same number of second-stage optimiser updates using only the labelled subset.}
\label{tab:pseudo_control}
\small
\begin{tabular*}{\textwidth}{@{}p{\textwidth}@{}}
\centering
\textit{Panel A: Pseudo-label selection}
\par\vspace{2pt}

\begin{tabular}{@{}rrrr@{}}
\toprule
\multicolumn{1}{c}{\shortstack{Labels\\(\%)}} &
\multicolumn{1}{c}{\shortstack{Labelled\\images}} &
\multicolumn{1}{c}{\shortstack{Pseudo\\labels}} &
\multicolumn{1}{c}{\shortstack{Pseudo-label\\coverage}} \\
\midrule
1 & 100 & $3418 \pm 587$ & $0.345 \pm 0.059$ \\
3 & 299 & $4260 \pm 215$ & $0.439 \pm 0.022$ \\
5 & 500 & $5122 \pm 311$ & $0.539 \pm 0.033$ \\
10 & 1000 & $5207 \pm 105$ & $0.579 \pm 0.012$ \\
20 & 2001 & $5051 \pm 139$ & $0.631 \pm 0.017$ \\
40 & 3999 & $4000 \pm 73$  & $0.666 \pm 0.012$ \\
\bottomrule
\end{tabular}

\par\vspace{8pt}

\raggedright
\textit{Panel B: Test performance}
\par\vspace{2pt}

\centering
\begin{tabular}{@{}rrrrr@{}}
\toprule
\multicolumn{1}{c}{\shortstack{Labels\\(\%)}} &
\multicolumn{1}{c}{\shortstack{Stage 1\\accuracy}} &
\multicolumn{1}{c}{\shortstack{Control\\accuracy}} &
\multicolumn{1}{c}{\shortstack{Semi-supervised\\accuracy}} &
\multicolumn{1}{c}{\shortstack{Semi-supervised\\$-$ control}} \\
\midrule
1 & $0.6349 \pm 0.0178$ & $0.6495 \pm 0.0125$ & $0.6462 \pm 0.0162$ & $-0.0033 \pm 0.0062$ \\
3 & $0.6995 \pm 0.0060$ & $0.7007 \pm 0.0080$ & $0.6942 \pm 0.0041$ & $-0.0065 \pm 0.0050$ \\
5  & $0.7078 \pm 0.0117$ & $0.7094 \pm 0.0122$ & $0.7059 \pm 0.0132$ & $-0.0036 \pm 0.0029$ \\
10 & $0.7145 \pm 0.0138$ & $0.7182 \pm 0.0119$ & $0.7182 \pm 0.0102$ & $0.0000 \pm 0.0100$ \\
20 & $0.7237 \pm 0.0035$ & $0.7278 \pm 0.0042$ & $0.7290 \pm 0.0061$ & $0.0012 \pm 0.0088$ \\
40 & $0.7298 \pm 0.0024$ & $0.7350 \pm 0.0034$ & $0.7332 \pm 0.0047$ & $-0.0018 \pm 0.0039$ \\
\bottomrule
\end{tabular}

\end{tabular*}

\end{table}

\section{Class-relative pseudo-label selection}
\label{app:pseudo_label_distribution}

Table~\ref{tab:pseudo_label_class_relative} reports class-relative pseudo-label selection diagnostics for the high-confidence pseudo-labeling experiment at threshold $\tau=0.95$. These diagnostics account for the different class sizes in the unlabelled pool. The accepted true count is the number of unlabelled samples from a given true class that passed the confidence threshold. True-class coverage is the fraction of unlabelled samples from that true class that were accepted as pseudo-labels. Accepted enrichment compares the accepted true-class fraction with the corresponding class fraction in the unlabelled pool. Predicted-class precision is the pooled precision among accepted pseudo-labels assigned to the corresponding predicted class.

\begin{table}[htbp]
\centering
\caption{\textbf{Class-relative pseudo-label selection diagnostics at threshold $\tau=0.95$.} Accepted true counts are reported as mean $\pm$ standard deviation over five random seeds, with zero-selection seeds included. True-class coverage and accepted enrichment are computed from pooled counts across seeds. Predicted-class precision is the pooled precision among accepted pseudo-labels assigned to each class.}
\label{tab:pseudo_label_class_relative}
\scriptsize
\setlength{\tabcolsep}{4pt}
\renewcommand{\arraystretch}{1.1}

\begin{tabular}{@{}rlrrrr@{}}
\toprule
\multicolumn{1}{c}{Labels (\%)} &
\multicolumn{1}{c}{Class} &
\multicolumn{1}{c}{\shortstack{Accepted true\\count}} &
\multicolumn{1}{c}{\shortstack{True-class\\coverage}} &
\multicolumn{1}{c}{\shortstack{Accepted\\enrichment}} &
\multicolumn{1}{c}{\shortstack{Predicted-class\\precision}} \\
\midrule
1 & Altocumulus & $309 \pm 69$ & $0.430$ & $1.25$ & $0.988$ \\
1 & Cirrus & $104 \pm 60$ & $0.091$ & $0.26$ & $0.943$ \\
1 & Clear sky & $1521 \pm 101$ & $0.715$ & $2.07$ & $0.993$ \\
1 & Cumulonimbus & $962 \pm 317$ & $0.324$ & $0.94$ & $0.929$ \\
1 & Cumulus & $174 \pm 103$ & $0.226$ & $0.66$ & $0.947$ \\
1 & Mixed & $15 \pm 8$ & $0.043$ & $0.12$ & $0.333$ \\
1 & Stratocumulus & $334 \pm 277$ & $0.182$ & $0.53$ & $0.734$ \\
3 & Altocumulus & $433 \pm 22$ & $0.617$ & $1.40$ & $0.986$ \\
3 & Cirrus & $337 \pm 69$ & $0.301$ & $0.69$ & $0.980$ \\
3 & Clear sky & $1695 \pm 115$ & $0.812$ & $1.85$ & $0.995$ \\
3 & Cumulonimbus & $1217 \pm 189$ & $0.418$ & $0.95$ & $0.953$ \\
3 & Cumulus & $271 \pm 102$ & $0.360$ & $0.82$ & $0.984$ \\
3 & Mixed & $21 \pm 11$ & $0.062$ & $0.14$ & $0.690$ \\
3 & Stratocumulus & $287 \pm 141$ & $0.160$ & $0.36$ & $0.897$ \\
5 & Altocumulus & $510 \pm 38$ & $0.740$ & $1.37$ & $0.979$ \\
5 & Cirrus & $414 \pm 50$ & $0.378$ & $0.70$ & $0.975$ \\
5 & Clear sky & $1763 \pm 59$ & $0.863$ & $1.60$ & $0.994$ \\
5 & Cumulonimbus & $1469 \pm 226$ & $0.515$ & $0.96$ & $0.940$ \\
5 & Cumulus & $428 \pm 98$ & $0.582$ & $1.08$ & $0.976$ \\
5 & Mixed & $34 \pm 11$ & $0.103$ & $0.19$ & $0.836$ \\
5 & Stratocumulus & $504 \pm 146$ & $0.287$ & $0.53$ & $0.910$ \\
10 & Altocumulus & $500 \pm 23$ & $0.766$ & $1.32$ & $0.984$ \\
10 & Cirrus & $516 \pm 56$ & $0.497$ & $0.86$ & $0.980$ \\
10 & Clear sky & $1768 \pm 33$ & $0.914$ & $1.58$ & $0.992$ \\
10 & Cumulonimbus & $1418 \pm 123$ & $0.525$ & $0.91$ & $0.963$ \\
10 & Cumulus & $511 \pm 44$ & $0.733$ & $1.27$ & $0.979$ \\
10 & Mixed & $50 \pm 17$ & $0.159$ & $0.28$ & $0.874$ \\
10 & Stratocumulus & $445 \pm 126$ & $0.268$ & $0.46$ & $0.939$ \\
20 & Altocumulus & $459 \pm 47$ & $0.791$ & $1.25$ & $0.987$ \\
20 & Cirrus & $552 \pm 26$ & $0.599$ & $0.95$ & $0.981$ \\
20 & Clear sky & $1606 \pm 35$ & $0.933$ & $1.48$ & $0.992$ \\
20 & Cumulonimbus & $1345 \pm 189$ & $0.560$ & $0.89$ & $0.959$ \\
20 & Cumulus & $499 \pm 32$ & $0.805$ & $1.27$ & $0.982$ \\
20 & Mixed & $82 \pm 18$ & $0.294$ & $0.46$ & $0.925$ \\
20 & Stratocumulus & $509 \pm 135$ & $0.345$ & $0.55$ & $0.928$ \\
40 & Altocumulus & $385 \pm 14$ & $0.885$ & $1.33$ & $0.979$ \\
40 & Cirrus & $450 \pm 49$ & $0.650$ & $0.98$ & $0.985$ \\
40 & Clear sky & $1210 \pm 24$ & $0.938$ & $1.41$ & $0.993$ \\
40 & Cumulonimbus & $1135 \pm 80$ & $0.630$ & $0.94$ & $0.962$ \\
40 & Cumulus & $411 \pm 18$ & $0.884$ & $1.33$ & $0.980$ \\
40 & Mixed & $47 \pm 8$ & $0.227$ & $0.34$ & $0.972$ \\
40 & Stratocumulus & $362 \pm 101$ & $0.327$ & $0.49$ & $0.955$ \\
\bottomrule
\end{tabular}

\vspace{2pt}

\footnotesize
\emph{Note:} Shortened labels correspond to the GCD sky-type groups defined in Section~\ref{sec:dataset}.
\end{table}

\section{Active-learning queried-class distribution and enrichment}
\label{app:active_query_distribution}

Table~\ref{tab:active_query_distribution} reports the sky-type-group composition of samples acquired by uncertainty-based active learning at each acquisition step. Queried counts, queried fractions, pre-acquisition pool fractions, and enrichment ratios are reported as mean $\pm$ standard deviation over five random seeds. The queried fraction denotes the proportion of the acquired batch belonging to each sky-type group. The pool fraction denotes the proportion of the remaining unlabelled pool belonging to that sky-type group immediately before the acquisition step. Enrichment is defined as the ratio between the queried fraction and this pre-acquisition pool fraction. Values above one indicate that a sky-type group was over-represented among queried samples relative to its availability in the current unlabelled pool, while values below one indicate under-representation. 

Zero-count classes are included when computing the means and standard deviations. This table complements the active-learning diagnostics in Section~\ref{sec:results_active_queries} by showing both the absolute number and relative fraction of queried samples from each sky-type group.

\begin{table}[htbp]
\centering
\caption{\textbf{Sky-type-group distribution and enrichment of samples queried by uncertainty-based active learning.} Queried counts, queried fractions, pre-acquisition pool fractions, and enrichment ratios are reported as mean $\pm$ standard deviation over five random seeds.}
\label{tab:active_query_distribution}
\scriptsize
\setlength{\tabcolsep}{3.5pt}
\renewcommand{\arraystretch}{1.1}

\begin{tabular}{@{}llrrrr@{}}
\toprule
\multicolumn{1}{c}{\shortstack{Acquisition\\step}} &
\multicolumn{1}{c}{Class} &
\multicolumn{1}{c}{\shortstack{Queried\\count}} &
\multicolumn{1}{c}{\shortstack{Queried\\fraction}} &
\multicolumn{1}{c}{\shortstack{Pool\\fraction}} &
\multicolumn{1}{c}{Enrichment} \\
\midrule
$1\to3\%$ & Cumulus & $17 \pm 13$ & $0.084 \pm 0.066$ & $0.077 \pm 0.000$ & $1.09 \pm 0.86$ \\
$1\to3\%$ & Altocumulus & $28 \pm 11$ & $0.141 \pm 0.054$ & $0.073 \pm 0.000$ & $1.94 \pm 0.74$ \\
$1\to3\%$ & Cirrus & $60 \pm 18$ & $0.301 \pm 0.092$ & $0.115 \pm 0.000$ & $2.61 \pm 0.80$ \\
$1\to3\%$ & Clear sky & $5 \pm 3$ & $0.025 \pm 0.016$ & $0.215 \pm 0.000$ & $0.12 \pm 0.08$ \\
$1\to3\%$ & Stratocumulus & $27 \pm 9$ & $0.138 \pm 0.045$ & $0.185 \pm 0.000$ & $0.75 \pm 0.24$ \\
$1\to3\%$ & Cumulonimbus & $19 \pm 7$ & $0.093 \pm 0.034$ & $0.300 \pm 0.000$ & $0.31 \pm 0.11$ \\
$1\to3\%$ & Mixed & $43 \pm 9$ & $0.218 \pm 0.043$ & $0.035 \pm 0.000$ & $6.26 \pm 1.23$ \\
\midrule
$3\to5\%$ & Cumulus & $31 \pm 13$ & $0.153 \pm 0.063$ & $0.077 \pm 0.001$ & $1.97 \pm 0.80$ \\
$3\to5\%$ & Altocumulus & $11 \pm 6$ & $0.056 \pm 0.028$ & $0.071 \pm 0.001$ & $0.78 \pm 0.40$ \\
$3\to5\%$ & Cirrus & $45 \pm 11$ & $0.225 \pm 0.057$ & $0.111 \pm 0.002$ & $2.01 \pm 0.48$ \\
$3\to5\%$ & Clear sky & $19 \pm 6$ & $0.094 \pm 0.030$ & $0.219 \pm 0.000$ & $0.43 \pm 0.14$ \\
$3\to5\%$ & Stratocumulus & $49 \pm 9$ & $0.242 \pm 0.045$ & $0.186 \pm 0.001$ & $1.30 \pm 0.24$ \\
$3\to5\%$ & Cumulonimbus & $28 \pm 11$ & $0.140 \pm 0.053$ & $0.305 \pm 0.001$ & $0.46 \pm 0.17$ \\
$3\to5\%$ & Mixed & $18 \pm 9$ & $0.091 \pm 0.045$ & $0.031 \pm 0.001$ & $2.91 \pm 1.44$ \\
\midrule
$5\to10\%$ & Cumulus & $30 \pm 13$ & $0.060 \pm 0.026$ & $0.076 \pm 0.001$ & $0.79 \pm 0.35$ \\
$5\to10\%$ & Altocumulus & $19 \pm 6$ & $0.039 \pm 0.012$ & $0.071 \pm 0.001$ & $0.54 \pm 0.17$ \\
$5\to10\%$ & Cirrus & $82 \pm 12$ & $0.164 \pm 0.025$ & $0.109 \pm 0.001$ & $1.51 \pm 0.22$ \\
$5\to10\%$ & Clear sky & $31 \pm 7$ & $0.062 \pm 0.015$ & $0.221 \pm 0.001$ & $0.28 \pm 0.07$ \\
$5\to10\%$ & Stratocumulus & $117 \pm 21$ & $0.233 \pm 0.041$ & $0.184 \pm 0.001$ & $1.27 \pm 0.23$ \\
$5\to10\%$ & Cumulonimbus & $173 \pm 31$ & $0.346 \pm 0.062$ & $0.308 \pm 0.001$ & $1.12 \pm 0.20$ \\
$5\to10\%$ & Mixed & $48 \pm 11$ & $0.096 \pm 0.023$ & $0.030 \pm 0.001$ & $3.19 \pm 0.64$ \\
\midrule
$10\to20\%$ & Cumulus & $59 \pm 17$ & $0.059 \pm 0.017$ & $0.077 \pm 0.002$ & $0.77 \pm 0.21$ \\
$10\to20\%$ & Altocumulus & $33 \pm 10$ & $0.033 \pm 0.010$ & $0.073 \pm 0.001$ & $0.45 \pm 0.13$ \\
$10\to20\%$ & Cirrus & $150 \pm 33$ & $0.150 \pm 0.033$ & $0.106 \pm 0.001$ & $1.41 \pm 0.30$ \\
$10\to20\%$ & Clear sky & $29 \pm 15$ & $0.029 \pm 0.015$ & $0.230 \pm 0.000$ & $0.13 \pm 0.06$ \\
$10\to20\%$ & Stratocumulus & $298 \pm 37$ & $0.297 \pm 0.037$ & $0.182 \pm 0.003$ & $1.64 \pm 0.21$ \\
$10\to20\%$ & Cumulonimbus & $368 \pm 33$ & $0.367 \pm 0.033$ & $0.306 \pm 0.003$ & $1.20 \pm 0.11$ \\
$10\to20\%$ & Mixed & $65 \pm 16$ & $0.065 \pm 0.016$ & $0.026 \pm 0.001$ & $2.47 \pm 0.57$ \\
\midrule
$20\to40\%$ & Cumulus & $110 \pm 47$ & $0.055 \pm 0.024$ & $0.079 \pm 0.003$ & $0.69 \pm 0.27$ \\
$20\to40\%$ & Altocumulus & $48 \pm 14$ & $0.024 \pm 0.007$ & $0.078 \pm 0.001$ & $0.31 \pm 0.09$ \\
$20\to40\%$ & Cirrus & $167 \pm 16$ & $0.083 \pm 0.008$ & $0.100 \pm 0.004$ & $0.83 \pm 0.09$ \\
$20\to40\%$ & Clear sky & $117 \pm 41$ & $0.058 \pm 0.020$ & $0.256 \pm 0.002$ & $0.23 \pm 0.08$ \\
$20\to40\%$ & Stratocumulus & $719 \pm 41$ & $0.360 \pm 0.020$ & $0.167 \pm 0.007$ & $2.15 \pm 0.10$ \\
$20\to40\%$ & Cumulonimbus & $729 \pm 37$ & $0.365 \pm 0.018$ & $0.298 \pm 0.004$ & $1.22 \pm 0.07$ \\
$20\to40\%$ & Mixed & $109 \pm 15$ & $0.055 \pm 0.008$ & $0.021 \pm 0.002$ & $2.55 \pm 0.21$ \\
\bottomrule
\end{tabular}

\vspace{2pt}

\footnotesize
\emph{Note:} Shortened labels correspond to the GCD sky-type groups defined in Section~\ref{sec:dataset}.

\end{table}

\end{appendices}


\bibliography{references}

\end{document}